\documentclass{article}

\PassOptionsToPackage{numbers, compress}{natbib}
\usepackage[final]{neurips_2022}

\usepackage[utf8]{inputenc} % allow utf-8 input
\usepackage[T1]{fontenc}    % use 8-bit T1 fonts
\usepackage{hyperref}       % hyperlinks
\usepackage{url}            % simple URL typesetting
\usepackage{booktabs}       % professional-quality tables
\usepackage{amsfonts}       % blackboard math symbols
\usepackage{nicefrac}       % compact symbols for 1/2, etc.
\usepackage{microtype}      % microtypography
\usepackage{xcolor}         % colors
\usepackage{colortbl}

\usepackage{amsmath}
\usepackage{amssymb}
\usepackage{mathtools}
\usepackage{amsthm}
\usepackage{subcaption}
\usepackage{enumitem}
\usepackage{multirow}
\usepackage{capt-of}
\usepackage{wrapfig}
\usepackage{makecell}
\usepackage{bm}
\usepackage{bbm}
\usepackage{arydshln}
\usepackage{placeins}
\usepackage{comment}
\usepackage{algorithm}
\usepackage{algorithmic}
\usepackage{wrapfig}
\usepackage{mathtools,eqparbox}
\usepackage{adjustbox}
\usepackage[capitalize,noabbrev]{cleveref}

\definecolor{gg}{gray}{0.92}
\newcolumntype{a}{>{\columncolor{gg}}c}

\definecolor{Red}{rgb}{0.6,0,0}
\definecolor{Blue}{rgb}{0,0,0.8}
\definecolor{Green}{rgb}{0,0.4,0.7}
\definecolor{airforceblue}{rgb}{0.36, 0.54, 0.66}
\definecolor{ao(english)}{rgb}{0.0, 0.5, 0.0}
\definecolor{azure(colorwheel)}{rgb}{0.0, 0.5, 1.0}
\definecolor{crimson}{rgb}{0.86, 0.08, 0.24}
\definecolor{darkcerulean}{rgb}{0.03, 0.27, 0.49}
\definecolor{cobalt}{rgb}{0.0, 0.28, 0.67}
\definecolor{rosegold}{rgb}{0.72, 0.43, 0.47}
\definecolor{orange-red}{rgb}{1.0, 0.27, 0.0}
\definecolor{mountainmeadow}{rgb}{0.19, 0.73, 0.56}
\definecolor{malachite}{rgb}{0.04, 0.85, 0.32}
\definecolor{darkblue}{rgb}{0.0, 0.0, 0.55}
\definecolor{customblue}{rgb}{0.2, 0.35, 0.8}

\hypersetup{
    colorlinks = true,
    citecolor = Blue,
    linkcolor = Red
}

\title{Graph Self-supervised Learning \\ with Accurate Discrepancy Learning}

\author{
  Dongki Kim$^{1}\thanks{Equal Contribution.}$,\quad Jinheon Baek$^{1*}$,\quad Sung Ju Hwang$^{1,2}$ \\
  KAIST$^{1}$, AITRICS$^{2}$, South Korea \\
  \texttt{cleverki@kaist.ac.kr, jinheon.baek@kaist.ac.kr, sjhwang82@kaist.ac.kr}
}

\let\oldmaketitle\maketitle
\renewcommand{\maketitle}{\oldmaketitle\setcounter{footnote}{0}}

\begin{document}

\maketitle
\begin{abstract}
Self-supervised learning of graph neural networks (GNNs) aims to learn an accurate representation of the graphs in an unsupervised manner, to obtain transferable representations of them for diverse downstream tasks. Predictive learning and contrastive learning are the two most prevalent approaches for graph self-supervised learning. However, they have their own drawbacks. While the predictive learning methods can learn the contextual relationships between neighboring nodes and edges, they cannot learn global graph-level similarities. Contrastive learning, while it can learn global graph-level similarities, its objective to maximize the similarity between two differently perturbed graphs may result in representations that cannot discriminate two similar graphs with different properties. To tackle such limitations, we propose a framework that aims to learn the exact discrepancy between the original and the perturbed graphs, coined as \textit{Discrepancy-based Self-supervised LeArning} (D-SLA). Specifically, we create multiple perturbations of the given graph with varying degrees of similarity, and train the model to predict whether each graph is the original graph or the perturbed one. Moreover, we further aim to accurately capture the amount of discrepancy for each perturbed graph using the graph edit distance. We validate our D-SLA on various graph-related downstream tasks, including molecular property prediction, protein function prediction, and link prediction tasks, on which ours largely outperforms relevant baselines\footnote{Code is available at https://github.com/DongkiKim95/D-SLA}.
\end{abstract}
\section{Introduction}

A graph, consisting of nodes and edges, is a data structure that defines a relationship among objects. 
Recently, graph neural networks (GNNs)~\cite{GCN, GraphSAGE, GAT, GIN}, which aim to represent this structure with neural networks, have achieved great successes in modeling real-world graphs such as social networks~\cite{socialnetworkapplication}, knowledge graphs~\cite{gen}, biological networks~\cite{bionetworkapplication}, and molecular graphs~\cite{MARS}. However, it is extremely costly and time-consuming to annotate every label of the graphs. For example, labeling the properties of molecular graphs requires time-consuming laboratory experiments~\cite{pretraingnns}. 

Recently, various self-supervised learning methods for GNNs~\cite{pretraingnns, graphbert, GPTGNN, SUGAR} have been studied to overcome this issue of the lack of labeled graph data. The existing self-supervised learning methods can be classified into two categories: predictive learning and contrastive learning. Predictive learning methods~\cite{pretraingnns, GROVER, SuperGAT} aim to learn representations by predicting contexts of a graph (Figure~\ref{fig:concept}, (a-1)). However, predictive learning schemes are limited as they consider only the subgraphical semantics. 
Contrastive learning~\cite{DGI, Infograph, GraphCL, GraphCLAdaptive, JOAO, InfoGCL}, which is a popular self-supervised learning approach in the computer vision field~\cite{visioncontra1, visioncontra2, simCLR, MoCo}, can learn the global semantics by maximizing the similarity between instances (images) perturbed by semantic-invariant perturbations (e.g. color jittering or flipping). Graph contrastive learning methods also use a similar strategy, generating positive examples with edge perturbation, attribute masking, and subgraph sampling, and then maximizing the similarity between two differently perturbed graphs from the original graph (See Figure~\ref{fig:concept}, (b-1)).

\begin{figure*}[t]
    \centering
    \begin{minipage}{0.32\linewidth}
        \centering
        \centerline{\includegraphics[width=0.975\textwidth]{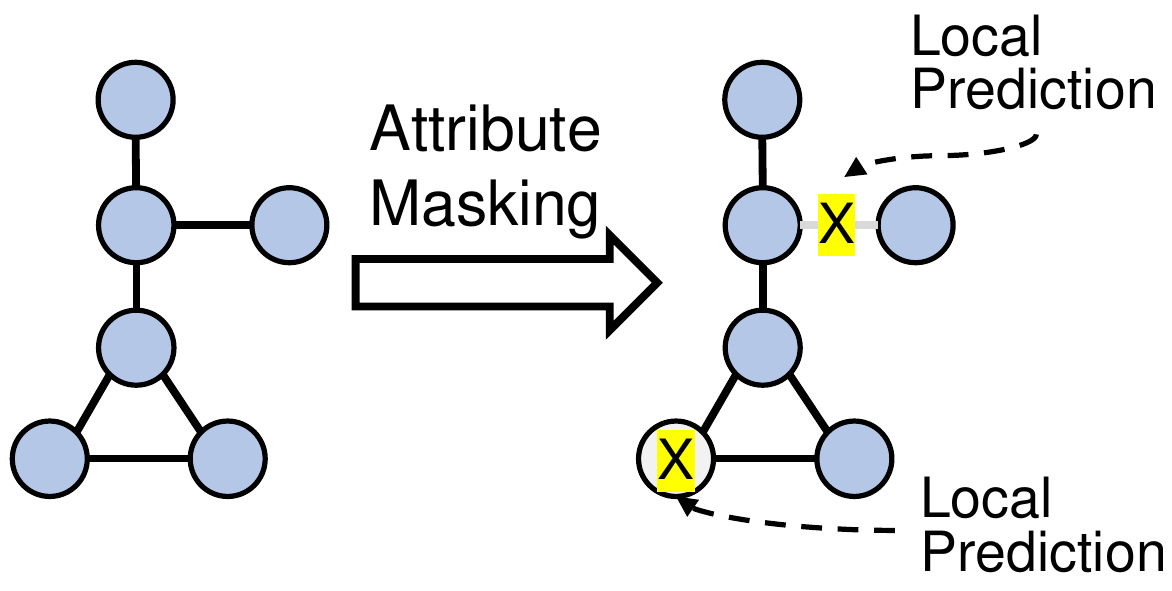}}
        \vspace{-0.05in}
        \subcaption*{(a-1) Conventional Predictive learning}
    \end{minipage}
    \hfill
    \begin{minipage}{0.32\linewidth}
        \centering
        \centerline{\includegraphics[width=0.975\textwidth]{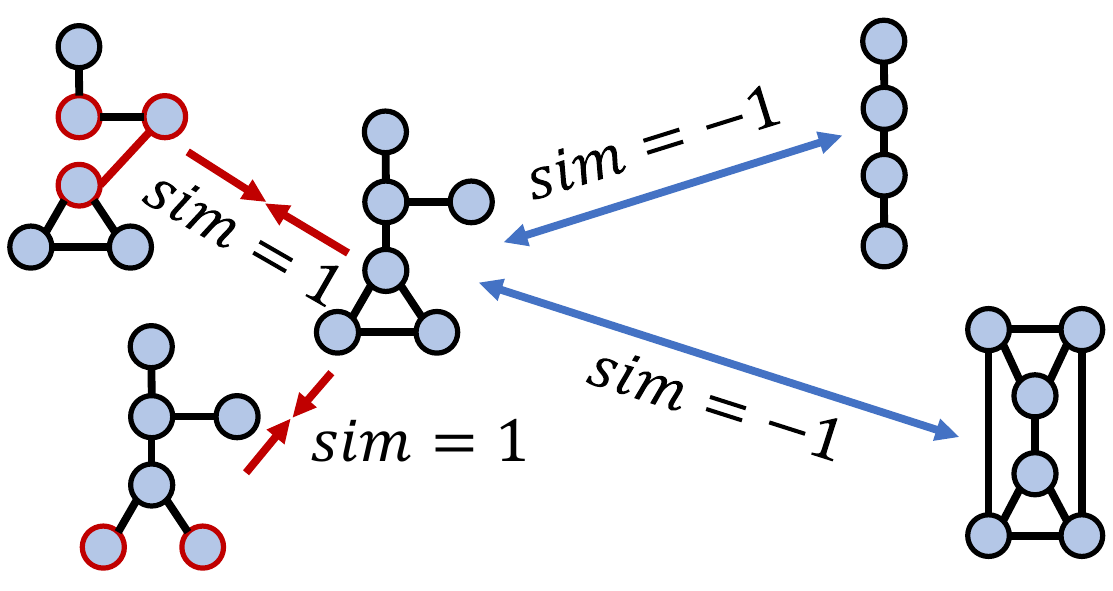}}
        \vspace{-0.1in}
        \subcaption*{(b-1) Conventional Contrastive learning}
    \end{minipage}
    \hfill
    \begin{minipage}{0.32\linewidth}
        \centering
        \centerline{\includegraphics[width=0.975\textwidth]{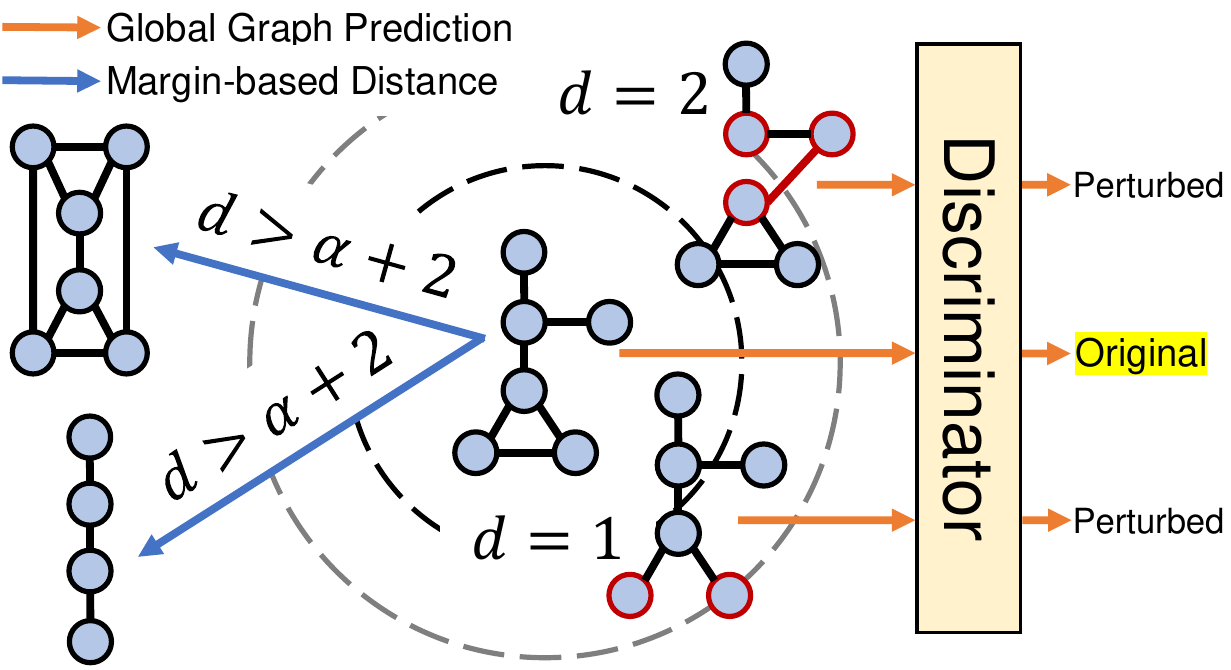}}
        \vspace{-0.1in}
        \subcaption*{(c-1) Our Discrepancy-based Learning}
    \end{minipage}
    \vspace{-0.17in}
\end{figure*}

\begin{figure*}[t]\ContinuedFloat
    
    \begin{minipage}{0.63\textwidth}
        \vspace{-0.1in}
        \begin{minipage}{0.32\textwidth}
        \centering
        \centerline{\includegraphics[width=0.975\linewidth]{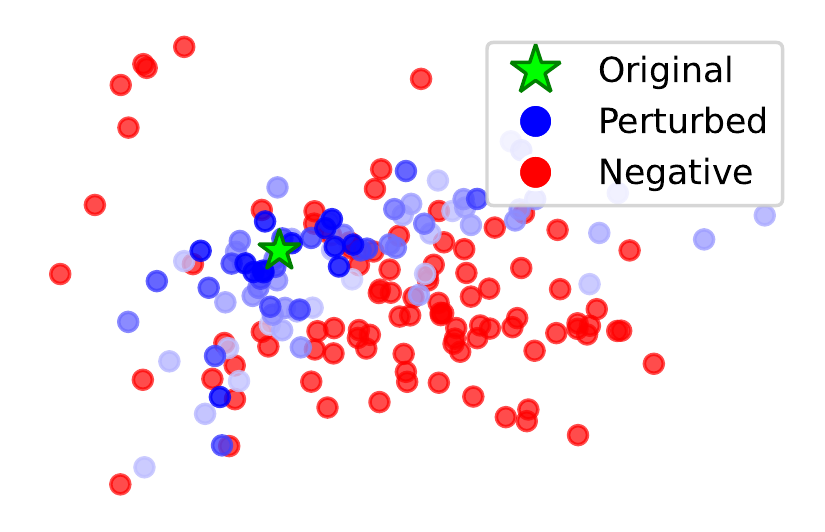}}
        \vspace{-0.1in}
        \subcaption*{(a-2) Predictive (AttrMask)}
        \end{minipage}
        \hfill
        \begin{minipage}{0.32\textwidth}
        \centering
        \centerline{\includegraphics[width=0.975\linewidth]{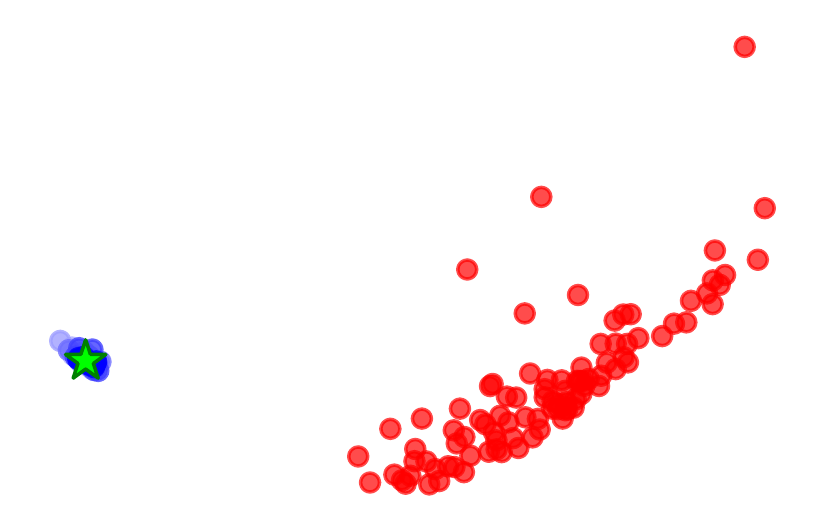}}
        \vspace{-0.1in}
        \subcaption*{(b-2) Contrastive (GraphCL)}
        \end{minipage}
        \hfill
        \begin{minipage}{0.32\textwidth}
        \centering
        \centerline{\includegraphics[width=0.975\linewidth]{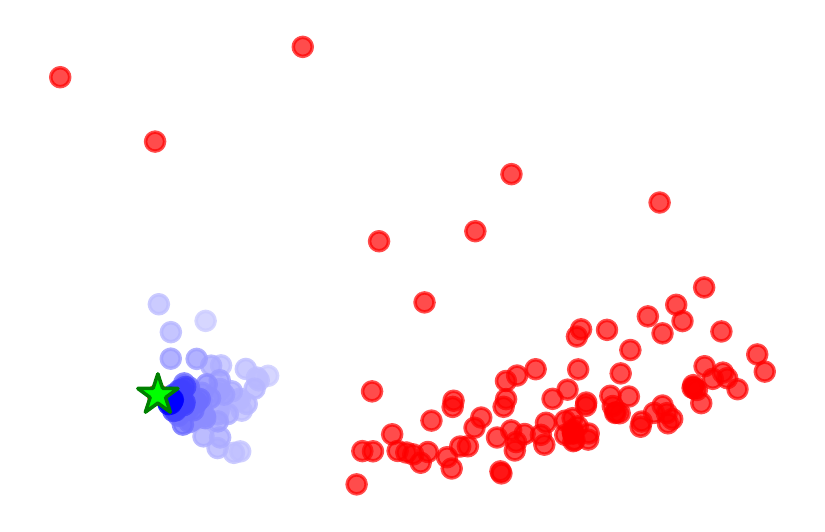}}
        \vspace{-0.1in}
        \subcaption*{(c-2) D-SLA (Ours)}
        \end{minipage}
        \vspace{-0.1in}
    \end{minipage}
    \hfill
    \begin{minipage}{0.35\textwidth}
        \centering
        \vspace{0.05in}
        \begin{minipage}{0.49\textwidth}
            \includegraphics[width=1\linewidth]{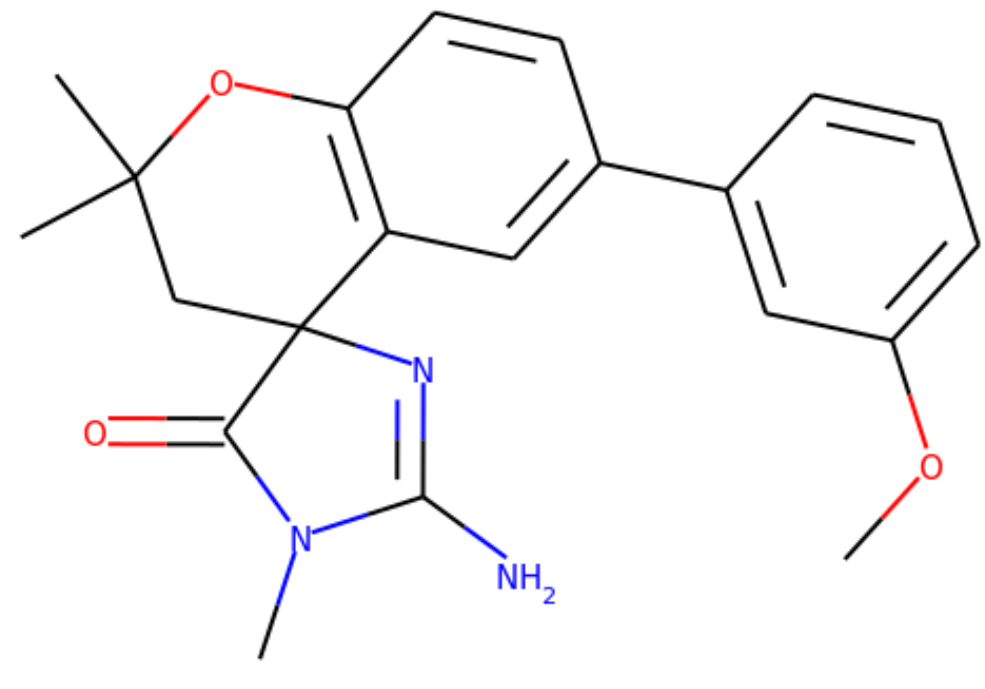}
            \vspace{-0.25in}
            \subcaption*{(d) Active}
        \end{minipage}
        \hfill
        \begin{minipage}{0.49\textwidth}
            \includegraphics[width=1\linewidth]{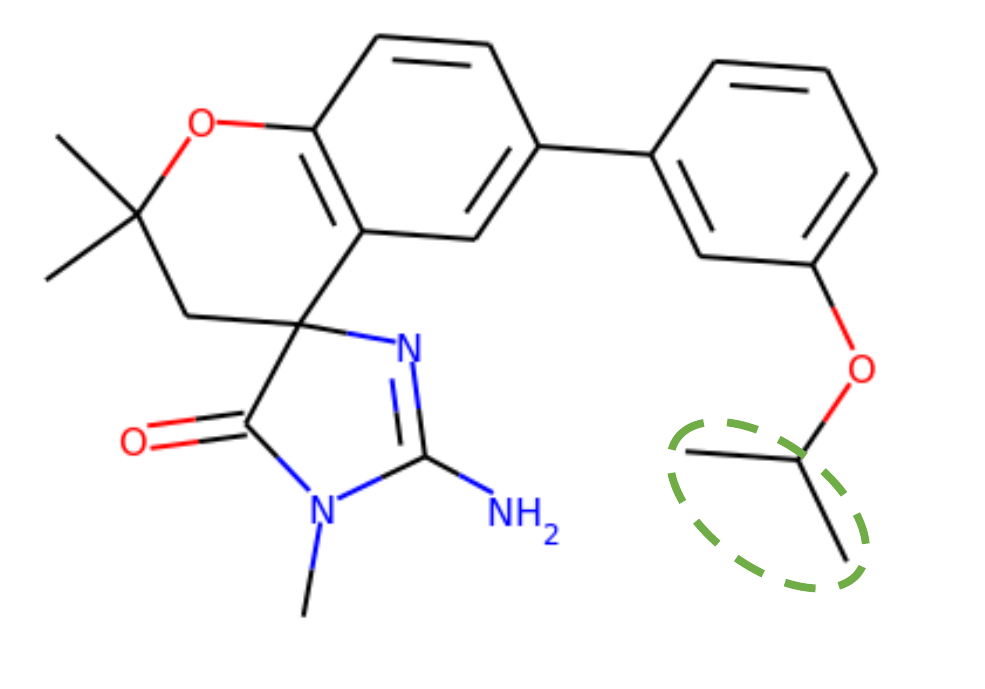}
            \vspace{-0.25in}
            \subcaption*{(e) Inactive}
        \end{minipage}
    \end{minipage}
    \vspace{-0.08in}
    \caption{\small \textbf{(a) Conventional predictive learning} that aims to predict local attributes by masking them. \textbf{(b) Conventional contrastive learning} that could maximize the similarity of dissimilar graphs perturbed from original graphs. \textbf{(c) Our discrepancy-based learning} that discriminates an original graph from perturbed graphs by capturing global semantics of them unlike predictive learning, but also that reflects the amount of discrepancy across original, perturbed, and different graphs, unlike contrastive learning. \textbf{(a,b,c-2) Graph embedding visualization} trained by each self-supervised learning scheme. \textbf{(d),(e) Motivation:}  A pitfall of contrastive learning methods, which assumes two similar graphs as the same, despite of their different properties.}
    \label{fig:concept}
    \vspace{-0.22in}
\end{figure*}

However, is it always correct to assume that the perturbed graphs by such perturbations will preserve the semantics of the original graph? In computer vision, there exists a set of well-defined perturbations, that does not change the human perception of the given image. However, in contrast to images that reside in a continuous domain, graphs are discrete data structures by nature, and thus their properties could become completely different even with slight perturbations. For example, two molecules in Figure~\ref{fig:concept} (d) and (e) show that, although they have a highly correlated structure, one molecular graph could be actively bound to a set of inhibitors of human $\beta$-secretase 1~\cite{moleculenet}, whereas the other is not. In this case, existing contrastive learning with perturbations may lead the representations for two semantically \textit{dissimilar} graphs to collapse into the one representation (See Figure~\ref{fig:concept} (b-2)).

To tackle this problem, we propose a novel self-supervised learning method which aims to learn the \textit{discrepancy} between graphs, which we refer to as \textbf{D}iscrepancy-based \textbf{S}elf-supervised \textbf{L}e\textbf{A}rning~(D-SLA). Specifically, we first perturb the given graph similarly as with contrastive learning schemes, but instead of maximizing the similarities across the perturbed graphs as done with contrastive learning, we aim to learn their discrepancies. To this end, we first introduce a discriminator which learns to discriminate real graphs from the perturbed ones (See Discriminator in Figure~\ref{fig:concept}, (c-1)). This allows the model to learn small differences that may largely impact the global property of the graph, as illustrated by the example in Figure~\ref{fig:concept} (d) and (e). However, simply knowing that two graphs are different is insufficient, and we would want to learn the exact amount of discrepancy between them. Notably, with our perturbation scheme, the exact discrepancy can be trivially obtained since the graph edit distance~\cite{grapheditdist} is automatically derived as the number of edges we perturb for a given graph. Then, we enforce the embedding-level difference between original and perturbed graphs to be proportional to the graph edit distance (See dotted circles in Figure~\ref{fig:concept} (c-1)). Finally, to discriminate the original and perturbed graphs from the completely different graphs, we enforce the relative distance of the latter to be larger than the former, by a large margin (See blue arrows in Figure~\ref{fig:concept}, (c-1)).

This will enable our D-SLA to capture both small and large topological differences (See Figure~\ref{fig:concept} (c-2)). Thus, our framework can enjoy the best of both worlds for graph self-supervised learning: predictive and contrastive learning. The experimental results show that our model significantly outperforms baselines, especially on the datasets where graphs with different properties have highly correlated structures, demonstrating the effectiveness of ours. Our main contributions are as follows:

\vspace{-0.12in}
\begin{itemize}[itemsep=0.4mm, parsep=1pt, leftmargin=*]
    \item We propose a novel graph self-supervised learning framework with a completely opposite objective from contrastive learning, which aims to learn to differentiate a graph and its perturbed ones using a discriminator, as even slight perturbations could lead to completely different properties for graphs. 
    
    \item Utilizing the graph edit distance that is obtained for perturbed graphs at no cost, we propose a novel objective to preserve the exact amount of discrepancy between graphs in the representation space.
    
    \item We validate our D-SLA by pre-training and fine-tuning it on various benchmarks of chemical, biology, and social domains, on which it significantly outperforms baselines.
\end{itemize}

\section{Related Work}
We now briefly review the existing works on graph neural networks (GNNs), and self-supervised learning methods for GNNs including predictive and contrastive learning.

\subsection{Graph Neural Networks}
\label{subsec:GNN}
\vspace{-0.05in}
Most existing graph neural networks (GNNs) could be formulated under the message passing scheme~\cite{MPNN}, which represents each node by firstly aggregating the features from its neighbors, and then combining the aggregated message with its own node representation. Different variants of update and aggregation functions have been studied. To mention a few, Graph Convolutional Network (GCN)~\cite{GCN} generalizes the convolution operation in a spectral domain of graphs, approximated by the mean aggregation. GraphSAGE~\cite{GraphSAGE} concatenates the representations of neighbors with its own representation when updating the node representation. Graph Attention Network (GAT)~\cite{GAT} considers the relative importance among neighboring nodes for neighborhood aggregation, which helps identify relevant neighbors for the given task. Graph Isomorphism Network (GIN)~\cite{GIN} uses the sum aggregation, allowing the model to distinguish two different graphs as powerfully as the Weisfeiler-Lehman (WL) test~\cite{WLtest}. While GNNs have achieved impressive results on various graph-related tasks, the trained representations from one dataset are usually not transferable to different downstream tasks. Moreover, labels of such task-oriented data are often scarce, especially in scientific domains (e.g., chemistry and biology)~\cite{pretraingnns}. Thus, in this work, we aim at obtaining transferable representations of graph-structured data with self-supervised learning, without using any labels. Self-supervised learning for GNNs can be broadly classified into two categories: predictive learning and contrastive learning, which we will briefly introduce in the following paragraphs.

\subsection{Predictive Learning for Graph Self-supervised Learning}
\vspace{-0.05in}
Predictive learning aims to learn contextual relationships by predicting subgraphical features, for example, nodes, edges, and subgraphs. Specifically, \citet{pretraingnns} propose to predict the attributes of masked nodes. Also, \citet{auxiliarySSL} and \citet{SuperGAT} propose to predict the presence of an edge or a path with the link prediction scheme. Furthermore, \citet{GPTGNN} and \citet{GROVER} propose to predict the generative sequence, contextual properties, and motifs of the given graphs. However, the predictive learning methods are limited in that they may not well capture the global structures and/or semantics of graphs.

\subsection{Contrastive Learning for Graph Self-supervised Learning}
\vspace{-0.05in}
The limitations of predictive learning gave rise to contrastive learning methods, which aim to capture global graph-level information. Early contrastive learning methods for GNNs aim to learn the similarity between the entire graph and its substructure, to obtain the representations for them without using any perturbations~\cite{DGI, Infograph}. Despite their successes, to further expand the embedding space of a model, contrastive learning methods have been actively studied with perturbation methods such as attribute masking, edge perturbation, and subgraph sampling~\cite{GraphCL, GraphCLAdaptive, MolCLR, JOAO}. Recently proposed adversarial methods further generate positive examples either by adaptively removing the edges~\cite{ADGCL} or by adjusting the attributes~\cite{GASSL}. Further, \citet{GraphLoG} aim to cluster graph representations via contrastive learning, and \citet{BGRL} propose a memory-efficient method for large-scale graph representation learning. However, while contrastive learning allows learning the global semantics with perturbed graphs, there exists a potential risk where the model can consider two dissimilar graphs as the same (See Figure~\ref{fig:concept} (b), (d), and (e)), due to the discrete nature of graphs. Contrarily, our novel pretext task of predicting the correct graph from multiple choices, which includes both slightly perturbed graphs and strongly perturbed graphs, allows the model to learn the correct relationships among local elements, but also learn the global graph-level differences.

\vspace{-0.05in}
\section{Method}
\begin{figure*}[t]
\centering
\centerline{\includegraphics[width=0.95\textwidth]{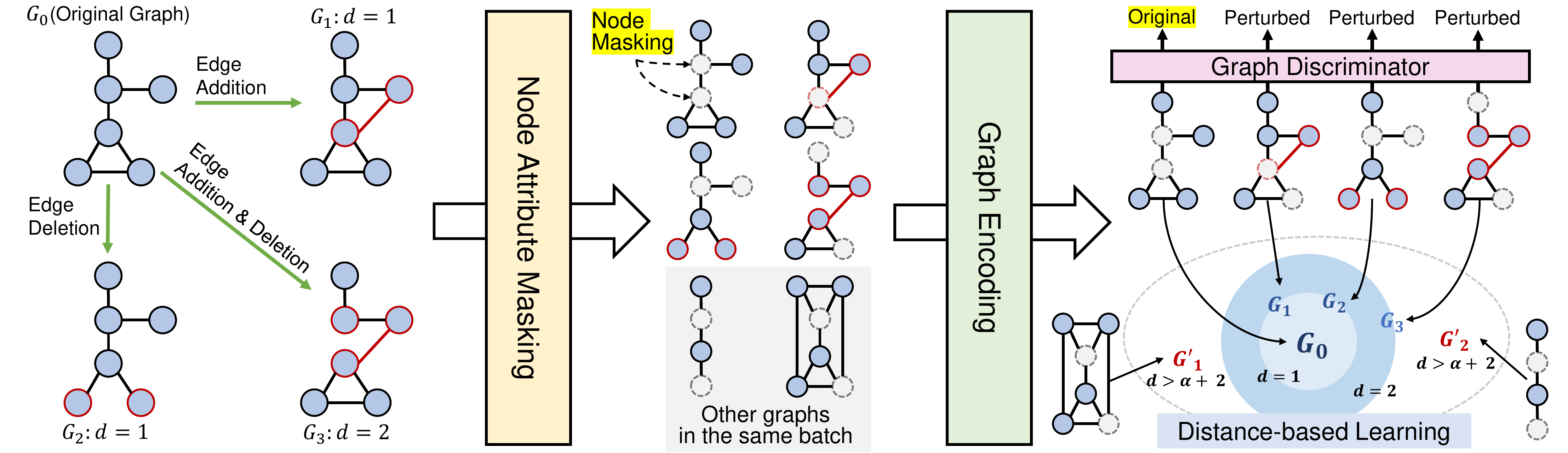}}
\vspace{-0.08in}
\caption{\small \textbf{Illustration of our overall framework (D-SLA).} We perturb the original graph by adding or deleting edges, where the discrepancy between original and perturbed graphs is defined by graph edit distance~\cite{grapheditdist}. Then, we mask the node attributes to make it more difficult to distinguish original and perturbed graphs. After that, we learn GNNs to distinguish perturbed graphs from original ones, but also accurately discriminate the original, perturbed, and other graphs on the embedding space with their distances.}
\vspace{-0.15in}
\label{fig:method}
\end{figure*}

\vspace{-0.05in}
In this section, we introduce our novel graph self-supervised learning framework, \textbf{D}iscrepancy-based \textbf{S}elf-supervised \textbf{L}e\textbf{A}rning~(\textit{D-SLA}), which is illustrated in Figure~\ref{fig:method}. Specifically, we first recount the notion of graph neural networks and existing graph self-supervised learning in section~\ref{subsec:preliminaries}. Then, we introduce each component of our D-SLA in Section~\ref{subsec:answerpred},~\ref{subsec:edit}, and~\ref{subsec:margin}, respectively. After that, we finally describe the overall framework that combines all three components in Section~\ref{subsec:overall}.

\subsection{Preliminaries}
\vspace{-0.05in}
\label{subsec:preliminaries}
\paragraph{Graph Neural Networks}
A graph $\mathcal{G}$ can be defined by four components: $\mathcal{G}=\left(\mathcal{V}, \mathcal{E}, \mathcal{X}_{\mathcal{V}}, \mathcal{X}_{\mathcal{E}}\right)$, where $\mathcal{V}$ is the set of nodes, $\mathcal{E}$ is the set of edges, $\mathcal{X}_\mathcal{V} \in \mathbb{R}^{|\mathcal{V}|\times d}$ is the matrix of node features, and $\mathcal{X}_\mathcal{E} \in \mathbb{R}^{|\mathcal{E}|\times d'}$ is the matrix of edge features, where $d$ and $d'$ are the dimensionalities of node and edge attributes, respectively. Let $h_v$ be a representation of a node $v$, and $h_\mathcal{G}$ be a representation of a graph $\mathcal{G}$. Then, the goal of graph neural networks (GNNs) is to represent nodes of the given graph by leveraging its topological structure. To be specific, GNNs learn node representations (i.e., $h_v$) by iteratively aggregating messages from their neighbors in a layer-wise manner, which is usually referred to as message-passing~\cite{MPNN} and formally defined as follows:
\begin{equation}
    h_v^{(l+1)} = f_U^{(l)}\left(h_v^{(l)}, f_A^{(l)}\left(\left\{h_u^{(l)}: \forall u \in \mathcal{N}(v)\right\}\right)\right),
    \label{eq:GNN}
\end{equation}
where $f_U^{(l)}$ denotes an update function that updates a representation of the given node along with its neighbors' representations, $f_A^{(l)}$ denotes an aggregate function that aggregates messages from the node's neighbors, $\mathcal{N}(v)$ denotes a set of neighbors of the node $v$, and $l$ denotes a $l$-th layer of GNNs. 

To further obtain the representation for an entire graph, we usually summarize all node representations from Equation~\ref{eq:GNN} into a single embedding vector with a permutation-invariant function $f_R$ as follows:
\begin{equation}
    h_\mathcal{G} = f_R\left(\left\{h_v^{(L)}: \forall v \in \mathcal{V}\right\}\right).
    \label{eq:readout}
\end{equation}
While a natural choice of $f_R$ is to use simple operations, such as mean or sum, various graph representation learning methods have been recently studied to accurately capture the entire graph information, including node clustering based~\cite{DiffPool, gmt}, and node drop based methods~\cite{TopKPool, SAGPool}.

\vspace{-0.05in}
\paragraph{Contrastive Learning for Self-supervised Learning of GNNs}
We now describe why contrastive learning fails to distinguish two topologically similar graphs yet having completely different properties. In particular, contrastive learning aims to increase the similarity between positive pairs of graphs while increasing the dissimilarity between negative pairs~\cite{GraphCL, GraphCLAdaptive, JOAO}. Thereat, to define positive pairs, they perturb the original graph by manipulating edges, masking attributes, and sampling subgraphs, and then consider the perturbed graphs as the same as the original graph. On the other hand, they consider other graphs in the same batch as dissimilar. Formally, for the perturbed graphs $\mathcal{G}_i$ and $\mathcal{G}_j$ from the original graph $\mathcal{G}_0$, the objective of contrastive learning is defined as follows:
\begin{equation}
    \mathcal{L}_{CL} = - \log \frac{f_{\text{sim}}(h_{\mathcal{G}_{i}}, h_{\mathcal{G}_{j}})}{\sum_{\mathcal{G}', \; \mathcal{G}' \ne \mathcal{G}_0} f_{\text{sim}}(h_{\mathcal{G}_{i}},  h_{\mathcal{G}'})},
    \label{eq:contra}
\end{equation}
where $\mathcal{G}'$ is the other graph in the same batch with the graph $\mathcal{G}_0$, which is also referred to as a \emph{negative graph}. Therefore, $\mathcal{G}_i$ and $\mathcal{G}_j$ are a positive pair, whereas, $\mathcal{G}_i$ and $\mathcal{G}'$ are a negative pair. $f_{\text{sim}}$ denotes a similarity function between two graphs, for example, $L_2$ distance or cosine similarity. By minimizing the objective in Equation~\ref{eq:contra}, existing contrastive learning methods closely embed two perturbed graphs $\mathcal{G}_i$ and $\mathcal{G}_j$: $\mathcal{G}_i \sim \mathcal{G}_j$. However, as the perturbations may not preserve the properties of the given graph due to the discrete nature of graph-structured data, they indeed should not be similar: $\mathcal{G}_i \not\sim \mathcal{G}_j$ (See Figure~\ref{fig:concept} (d) and (e)). To this end, we propose a new objective that can differentiate between the original graph and its perturbations, which we describe in the next subsection.

\subsection{Discrepancy Learning with Graph Discrimination}
\label{subsec:answerpred}
\vspace{-0.05in}
We now introduce our novel graph self-supervised learning objective to preserve the discrepancy between two graphs in the learned representation space, for which we then describe the perturbation scheme to generate augmented graphs from the original graph.

\vspace{-0.05in}
\paragraph{Discriminating Original Graphs from Perturbed Graphs} Contrarily to contrastive learning that trains perturbed graphs to be similar, our goal is to distinguish the original graph from perturbed graphs, by predicting the original graph among original and perturbed graphs, which we term as the original graph discrimination (See Graph Discriminator in Figure~\ref{fig:method}). Let $\mathcal{G}_0$ be an original graph, and $\mathcal{G}_i$ be a perturbed graph with $i \ge 1$. Then objective of our discrimination scheme is as follows:
\begin{equation}
    \mathcal{L}_{GD} = -\log\left(\frac{e^{S_{0}}}{e^{S_{0}} + \sum_{i\geq1}{e^{S_{i}}}}\right) \; \text{with} \; S = f_S(h_\mathcal{G}),
    \label{eq:answerprediction}
\end{equation}
where $S_k$ is a score of the graph $\mathcal{G}_k$, obtained from a learnable score network $f_S$ of the discriminator, i.e., $f_S: h_\mathcal{G} \mapsto S \in \mathbb{R}$. By training to discriminate the original graph from the perturbed graphs, the model is enforced to embed the perturbed graphs apart from the original graph (Figure \ref{fig:emb_ablation} (b)). Therefore, the model learned by our discrimination scheme will be capable of distinguishing even the slight differences between the original graph and its perturbations, as well as be capable of capturing the correct distribution of the graphs, since perturbed graphs could be semantically incorrect.

\vspace{-0.05in}
\paragraph{Perturbation}
The remaining question is how to perturb the graph to generate negative examples (i.e., $\mathcal{G}_i$). We aim at adding or deleting the \textit{subset} of edges in the given graph, and thus such perturbed graphs are slightly different from the original graph but also could be semantically incorrect. Again, in this way, since our graph discrimination objective in Equation~\ref{eq:answerprediction} discriminates the original graph from slightly perturbed graphs, the graph-level representations from our pre-trained GNNs can capture subtle difference that may have a large impact on the properties of graphs, for downstream tasks.

In particular, our perturbation scheme consists of the following two steps: 1) removing and adding a small number of edges, and 2) masking node attributes (Figure~\ref{fig:method}). To be specific, given an original graph $\mathcal{G}=\left(\mathcal{V}, \mathcal{E}, \mathcal{X}_{\mathcal{V}}, \mathcal{X}_{\mathcal{E}}\right)$, we aim to perturb it $n$ times, to obtain $\left\{ \mathcal{G}_1, ..., \mathcal{G}_n \right\}$. To do so, we first manipulate the edge set $\mathcal{E}$ by removing existing edges as well as adding new edges on it and then adjust its corresponding edge matrix $ \mathcal{X}_{\mathcal{E}}$. After that, we further randomly mask the node attributes on $\mathcal{X}_{\mathcal{V}}$ for both original and perturbed graphs, to make it more difficult to distinguish between them. Formally, the original and perturbed graphs from our node/edge perturbations are obtained as follows:
\begin{align}
    \mathcal{G}_0 \! &= \! \left(\mathcal{V}, \mathcal{E}, \tilde{\mathcal{X}}^0_{\mathcal{V}}, \mathcal{X}_{\mathcal{E}}\right)\! ,\: \tilde{\mathcal{X}}^0_{\mathcal{V}} \sim \texttt{M}(\mathcal{G}), \\
    \mathcal{G}_i \! &= \! \left(\mathcal{V}, \mathcal{E}^i, \tilde{\mathcal{X}}^i_{\mathcal{V}}, \mathcal{X}^i_{\mathcal{E}}\right)\! ,\: \tilde{\mathcal{X}}^i_{\mathcal{V}} \sim \texttt{M}(\mathcal{G}) ,\: (\mathcal{E}^i, \mathcal{X}^i_{\mathcal{E}}) \sim \texttt{P}(\mathcal{G}), \nonumber
\end{align}
where $\texttt{M}$ and $\texttt{P}$ are the node masking and edge perturbation functions, respectively. The scores of original and perturbed graphs are then computed with the score function $f_S$: $[S_0, S_1, ..., S_n] = [f_S(\mathcal{G}_0), f_S(\mathcal{G}_1), ..., f_S(\mathcal{G}_n)]$, which are then used for our previous learning objective in Equation~\ref{eq:answerprediction}.

\subsection{Learning Discrepancy with Edit Distance}
\label{subsec:edit}
\begin{figure*}[t]
    \begin{minipage}{0.245\textwidth}
    \centering
    \centerline{\includegraphics[width=0.975\linewidth]{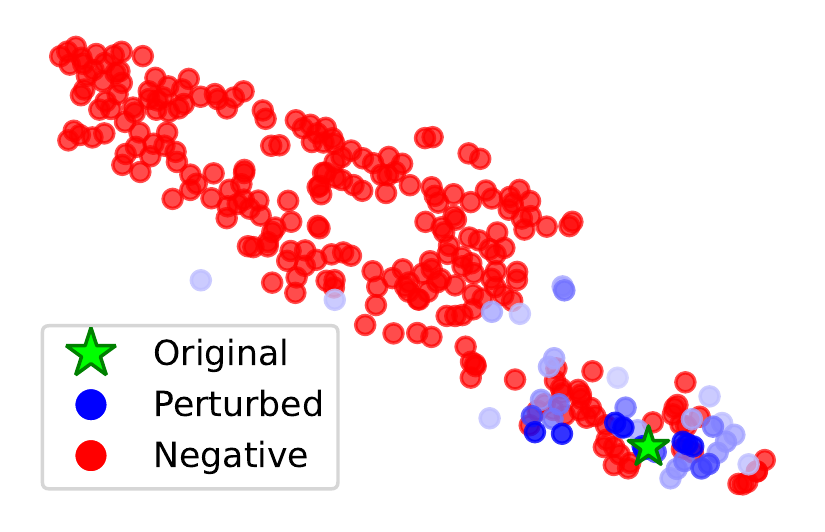}}
    \vspace{-0.08in}
    \subcaption{Randomly Init.}
    \end{minipage}
    \hfill
    \begin{minipage}{0.245\textwidth}
    \centering
    \centerline{\includegraphics[width=0.975\linewidth]{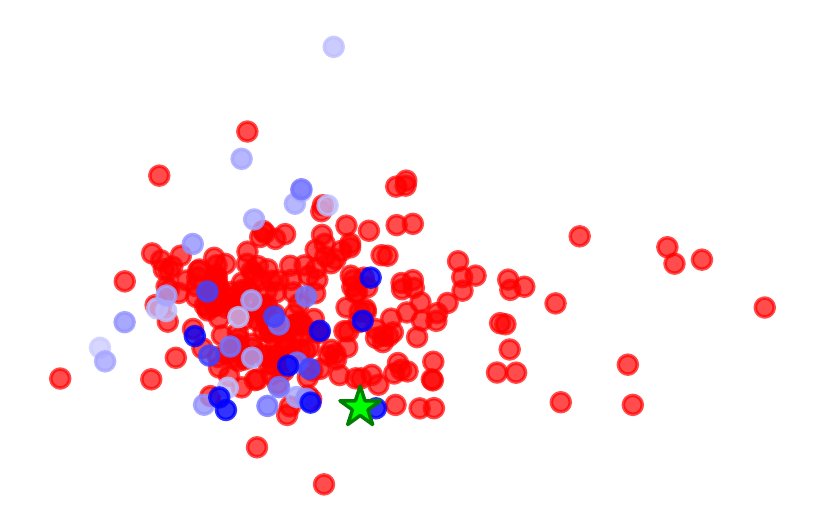}}
    \vspace{-0.08in}
    \subcaption{$\mathcal{L}_{GD}$}
    \end{minipage}
    \hfill
    \begin{minipage}{0.245\textwidth}
    \centering
    \centerline{\includegraphics[width=0.975\linewidth]{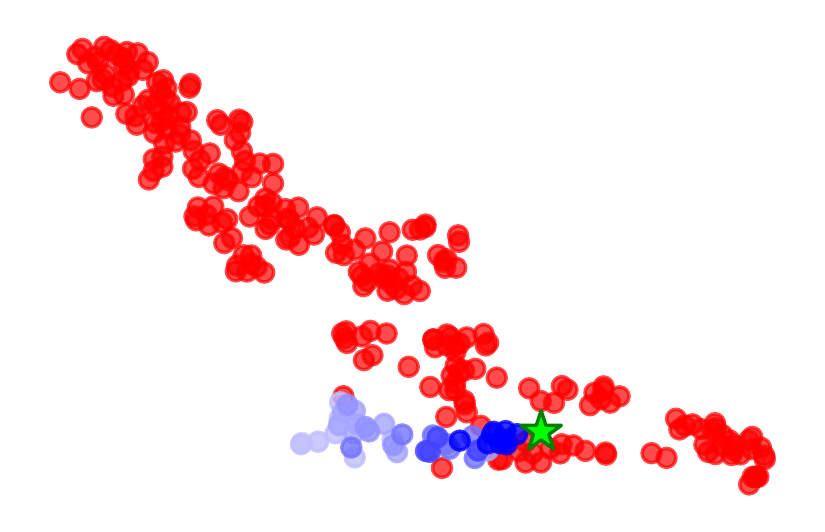}}
    \vspace{-0.08in}
    \subcaption{$\mathcal{L}_{GD}+\mathcal{L}_{edit}$}
    \end{minipage}
    \hfill
    \begin{minipage}{0.245\textwidth}
    \centering
    \centerline{\includegraphics[width=0.975\linewidth]{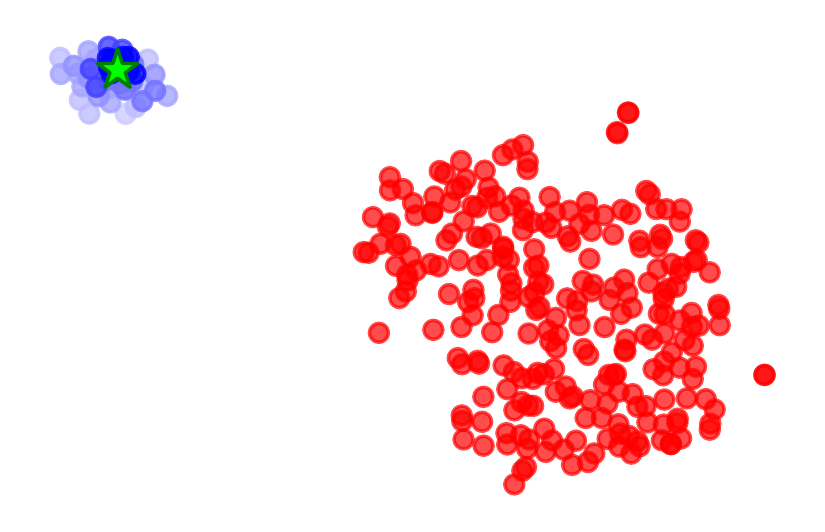}}
    \vspace{-0.08in}
    \subcaption{$\mathcal{L}_{GD}+\mathcal{L}_{edit}+\mathcal{L}_{margin}$}
    \end{minipage}
    \vspace{-0.05in}
    \caption{\small Visualizations of the graph embeddings to see the effect of each loss for our D-SLA. We represent the level of perturbations via the transparency of color, i.e., the stronger the perturbation, the lighter the color.
    }
    \label{fig:emb_ablation}
    \vspace{-0.175in}
\end{figure*}

\vspace{-0.05in}
While the learning to discriminate original and perturbed graphs described in Section~\ref{subsec:answerpred} allows the model to learn the discrepancy between original and perturbed graphs by embedding them onto different points, it cannot learn \textit{how dissimilar} the perturbed graph is from the original one as all perturbed graphs are considered as equal (Figure \ref{fig:emb_ablation} (b)). Thus, to further learn the accurate discrepancy among different graphs, here we introduce a method for efficiently computing the graph discrepancy between original and perturbed graphs leveraging the graph edit distance~\cite{grapheditdist}. Then we describe how to preserve the calculated discrepancy in the learned graph representation space.

\vspace{-0.05in}
\begin{wraptable}{t}{0.4\textwidth}
    \vspace{-0.18in}
    \small
    \centering
    \caption{\small Costs for calculating distance, with $m$ edges and $h$ iterations for WL-kernel~\cite{shervashidze2011weisfeiler}.}
    \vspace{-0.1in}
    \resizebox{0.39\textwidth}{!}{
    \renewcommand{\tabcolsep}{0.7mm}
    \begin{tabular}{lc}
    \toprule
    {\bf Methods} & {\bf Costs} \\
    \midrule
    Graph edit distance~\cite{grapheditdist} & NP-hard~\cite{edit/np-hard} \\
    Weisfeiler-Lehman kernel~\cite{shervashidze2011weisfeiler} & $\mathcal{O}(hm)$ \\
    Graph edit distance \textbf{with our perturbation} & $\mathcal{O}(1)$ \\
    \bottomrule
    \end{tabular}
    }
    \vspace{-0.1in}
    \label{tab:graphsim}
\end{wraptable}
\paragraph{Graph Edit Distance} To preserve the exact amount of discrepancies between original and perturbed graphs in the learned representation space, we first need to measure the graph distance. Graph edit distance, which is widely used to measure the dissimilarity between two graphs~\cite{grapheditdist}, is defined by the number of insertion, deletion, and substitution operations for nodes and edges, to transform one graph into the other graph. Although computing the graph edit distance is the NP-hard problem~\cite{edit/np-hard}, for our case, it is trivially obtained without any cost. This is because, for each perturbed graph, we know the exact number of edge addition and deletion steps as we have generated it using the same process. In other words, the number of added and deleted edges is simply the edit distance between original and perturbed graphs. This brings us significant computational advantages, as we generally require a significant amount of costs for calculating or estimating the distances between two graphs, as shown in Table~\ref{tab:graphsim}. Note that although several existing works~\cite{GMN, GraphSim} leverage graph edit distance, they are completely different from our work since they are neither self-supervised learning methods nor leverage the graph perturbation.

\vspace{-0.05in}
\paragraph{Distance-based Discrepancy Learning}
Based upon the graph edit distance, we design the regularization term to learn the exact amount of differences between original and perturbed graphs over the embedding space. To be specific, we regularize the model to learn that the embedding-level difference between original and perturbed graphs is proportional to their actual graph edit distance (i.e., if the edit distance between two graphs is large, then they are far away in the embedding space, whereas, if it is small, then they are close to each other) as shown in Distance-based Learning of Figure~\ref{fig:method}.

We first let an original graph $\mathcal{G}_0$ be an anchor graph, then a graph edit distance and an embedding-level distance between the anchor graph $\mathcal{G}_0$ and the perturbed graph $\mathcal{G}_i$ are defined as $e_i$ and $d_i$, respectively. With such notations, to achieve our objective of learning an exact amount of discrepancy across original and perturbed graphs, we formalize our edit distance-based regularization loss as follows:
\begin{equation}
    \mathcal{L}_{edit} = \sum\nolimits_{i, j} \left(\frac{d_i}{e_i} - \frac{d_j}{e_j}\right)^2 \; \text{with} \:\: d_i =  f_\text{diff}(h_{\mathcal{G}_0}, h_{\mathcal{G}_i}),
    \label{eq:editdistance}
\end{equation}
where $f_\text{diff}$ measures the embedding-level differences between graphs with $L_2$-norm. We find that Equation~\ref{eq:editdistance} can capture the exact amount of differences as shown in Figure \ref{fig:emb_ablation} (c). Note that Equation~\ref{eq:editdistance} alone does not work as shown in Table~\ref{tab:ablation}, since the trivial solution of Equation~\ref{eq:editdistance} is to set all the embedding-level distance between original and perturbed graphs as zero (i.e., $d_i=0 \; \forall i)$. However, learning the accurate amount of discrepancy without the trivial solution is feasible by jointly training with the objective of original graph discrimination in Equation~\ref{eq:answerprediction} where the model should differently embed $\mathcal{G}_0$ and $\mathcal{G}_i$ to discriminate them. Furthermore, as we can generate different levels of perturbations according to the graph edit distance, our graph embedding space can interpolate between two different graphs, for example, weakly and strongly perturbed graphs in terms of their actual distances, unlike previous predictive and contrastive learning methods.

\subsection{Relative Discrepancy Learning with Other Graphs}
\label{subsec:margin}
\vspace{-0.05in}
% , with the objectives of original graph discrimination
Thus far, we aimed toward distinguishing the perturbed graphs from the original graph in Section~\ref{subsec:answerpred}, with their actual edit distance-based discrepancy learning in Section~\ref{subsec:edit}. However, such schemes alone will not allow the model to learn the discrepancy between two completely different graphs (Figure \ref{fig:emb_ablation} (c)). While one might consider learning the discrepancy between original and its negative graphs (other graphs in the same batch) with their graph edit distance, it is impractical as computing the edit distance between completely different graphs is NP-hard~\cite{edit/np-hard}. Thus, we instead propose to learn a relative distance, exploiting the assumption that the distance between original and negative graphs in the same batch is larger than the distance between original and perturbed graphs with some amount of \textit{margin} (See Figure~\ref{fig:method}, Distance-based Learning). The usage of margin is highly beneficial for our discrepancy-based learning framework, since, if the negative graphs are far apart than the amount of margin plus the distance between original and perturbed graphs, the model does not attract the perturbed graphs to the original graph, therefore not losing the discrepancy learned in Equation~\ref{eq:editdistance}.

Formally, we realize the relative discrepancy learning with the triplet margin loss as follows:
\begin{equation}
    \mathcal{L}_{margin} = \sum\nolimits_{i, j} \max(0, \; \alpha + d_i - d'_j),
    \label{eq:margin}
\end{equation}
where $d_i$ is the distance between original and its perturbed graphs: $d_i = f_\text{diff}(h_{\mathcal{G}_0}, h_{\mathcal{G}_i})$, $d'_j$ is the distance between original and its negative graphs: $d'_j = f_\text{diff}(h_{\mathcal{G}_0}, h_{\mathcal{G}'_j})$ with $\mathcal{G}'$ as in-batch negative graphs, and $\alpha > 0$ as a margin hyperparameter. In this way, we can allow the model to learn that, for the original graph, negative graphs are more dissimilar than the perturbed graphs, while perturbed graphs are also marginally dissimilar to it with the amount of edit distances as shown in Figure~\ref{fig:emb_ablation} (d).

\subsection{Overall Framework}
\label{subsec:overall}
\vspace{-0.05in}
To sum up, our D-SLA framework aims to learn a graph representation space which preserves the discrepancy among graphs. Our graph discriminator described in Section~\ref{subsec:answerpred} helps learn a space that can discriminate even the slightly perturbed graph from its original graph (Figure~\ref{fig:emb_ablation} (b)). Then, the discrepancy learning with graph edit distance described in Section~\ref{subsec:edit} (Figure~\ref{fig:emb_ablation} (c)) enforces the embedding space to preserve the exact discrepancy among graphs. Finally, the margin-based triplet constraints between the perturbed and completely different graphs in Section~\ref{subsec:margin} allow the learned embedding space to capture the relative distance between graphs (Figure~\ref{fig:emb_ablation} (d)). Our overall learning objective, dubbed as Discrepancy-based Self-supervised LeArning (D-SLA), is given as follows:
\begin{equation}
    \mathcal{L} = \mathcal{L}_{GD} + \lambda_{1} \mathcal{L}_{edit} + \lambda_{2} \mathcal{L}_{margin},
    \label{eq:overall}
\end{equation}
where hyperparameters $\lambda_1$ and $\lambda_2$ are scaling weights to each loss. Note that, unlike predictive learning~\cite{pretraingnns} which aims to capture local semantics of graphs by masking and predicting local components, such as nodes and edges, our D-SLA learns the graph-level representations where the subtle differences on corner regions of different graphs could be captured by distinguishing original and perturbed graphs. Moreover, our D-SLA can discriminate differently perturbed graphs having different properties, unlike contrastive learning~\cite{GraphCL, JOAO} that considers them as similar.

\section{Experiments}
\vspace{-0.05in}
In this section, we first experimentally validate the proposed Discrepancy-based Self-supervised LeArning (D-SLA) on graph classification tasks to verify its effectiveness in obtaining accurate graph-level representations. After that, we further evaluate our D-SLA on link prediction tasks for which capturing the local semantics of graphs is important.

\subsection{Graph Classification}
\label{subsec:graph_classification}
\vspace{-0.05in}
Accurately capturing the global semantics of given graphs is crucial for graph classification tasks, on which we validate the performance of our D-SLA against existing baselines.

\vspace{-0.05in}
\paragraph{Experimental Setup}
We evaluate our D-SLA on two different domains: the molecular property prediction task from chemical domain~\cite{moleculenet, OGB} and the protein function prediction task from biological domain~\cite{pretraingnns}. 
For the chemical domain, we follow the experimental setup from~\citet{JOAO}, whose goal is to predict the molecules' biochemical activities. For the self-supervised learning, we use 2M molecules from the ZINC15 dataset~\cite{ZINC15}. Then, after the self-supervised learning, we perform fine-tuning on datasets from MoleculeNet~\cite{moleculenet} to evaluate the down-stream performances of models. 
For the biological domain, we follow the setup from~\citet{JOAO}, for which the goal is to predict the proteins' biological functions. For pre-training and fine-tuning, we use the dataset of PPI networks~\cite{PPI/pretrain}.
We use the ROC-AUC value as an evaluation metric, and report the average performance over five different runs following~\citet{GraphLoG}. See Appendix \ref{appen:graph_classification} for dataset details.

\vspace{-0.05in}
\paragraph{Models}
We compare our D-SLA against predictive learning baselines: EdgePred~\cite{GraphSAGE}, AttrMasking~\cite{pretraingnns} and ContextPred~\cite{pretraingnns}, and contrastive learning baselines: Infomax~\cite{DGI}, GraphCL~\cite{GraphCL}, JOAO~\cite{JOAO}, GraphLoG~\cite{GraphLoG}, and BGRL~\cite{BGRL}. Detailed explanations of models including baselines and ours are provided in Appendix \ref{appen:baselines}.

\vspace{-0.05in}
\paragraph{Implementation Details}
We follow the conventional experimental setup of graph self-supervised learning from~\citet{pretraingnns}, where we use the GIN~\cite{GIN} as the base network. For pre-training of our model, we perturb the original graph three times. To be specific, to obtain the perturbed graphs from the original graph, we first randomly select a subgraph of it, and then add or remove edges in the range of $\left\{ 20\%, 40\%, 60\% \right\}$ over the sampled subgraph. For fine-tuning, we follow the hyperparameters from~\citet{JOAO}. We provide additional details in Appendix \ref{appen:graph_classification}.

\vspace{-0.05in}
\paragraph{Results}
Table~\ref{tab:graph_classification} shows that our D-SLA achieves the best average performance against existing predictive and contrastive learning baselines on tasks from both chemical and biological domains, demonstrating the effectiveness of our discrepancy-based framework. To better see what aspects of D-SLA contribute to the performance improvements, we perform in-depth analyses of each dataset, and how our model can effectively handle the task on it, in the next paragraph.
%%%%%%%%%%%%%%%%%%%%%%%%%%%%%%%%%%%%%%%%%%%%%%%%%%%%%%%%%%%%%%%%%%%%%%%%%%%%%%%%%%%%%%
\begin{table*}[t]
    \caption{\small Fine-tuning results on graph classification tasks of chemical and biological domains. Best performances are highlighted in bold. The reported results are taken from \citet{JOAO} and \citet{GraphLoG}, except for the BGRL results and the PPI performance of GraphLoG as they are not available.
    }
    \vspace{-0.075in}
    \begin{adjustbox}{width=\textwidth}
    \renewcommand{\arraystretch}{1.2}
    \renewcommand{\tabcolsep}{2.0mm}
    \begin{tabular}{llccccccccca}
    \toprule
    & SSL methods & BBBP & ClinTox & MUV & HIV & BACE & SIDER & Tox21 & ToxCast & PPI & Avg. \\
    \midrule
    & No Pretrain & 65.8 $\pm$ 4.5 &   58.0 $\pm$ 4.4 &   71.8 $\pm$ 2.5 &   75.3 $\pm$ 1.9 &   70.1 $\pm$ 5.4 &   57.3 $\pm$ 1.6 &   74.0 $\pm$ 0.8 &   63.4 $\pm$ 0.6 &   64.8 $\pm$ 1.0 & 66.72 \\
    \midrule

    \multirow{3}{*}{\rotatebox{90}{\textit{\fontsize{8pt}{8pt}\selectfont Predictive}}}
    & Edgepred &    67.3 $\pm$ 2.4 &   64.1 $\pm$ 3.7 &   74.1 $\pm$ 2.1 &   76.3 $\pm$ 1.0 &   79.9 $\pm$ 0.9 &   60.4 $\pm$ 0.7 &   76.0 $\pm$ 0.6 &   64.1 $\pm$ 0.6 &   65.7 $\pm$ 1.3 & 69.77\\
    & AttrMaskig &  64.3 $\pm$ 2.8 &   71.8 $\pm$ 4.1 &   74.7 $\pm$ 1.4 &   77.2 $\pm$ 1.1 &   79.3 $\pm$ 1.6 &   61.0 $\pm$ 0.7 &   76.7 $\pm$ 0.4 &   64.2 $\pm$ 0.5 &   65.2 $\pm$ 1.6 & 70.49 \\
    & ContextPred & 68.0 $\pm$ 2.0 &   65.9 $\pm$ 3.8 &   75.8 $\pm$ 1.7 &   77.3 $\pm$ 1.0 &   79.6 $\pm$ 1.2 &   60.9 $\pm$ 0.6 &   75.7 $\pm$ 0.7 &   63.9 $\pm$ 0.6 &   64.4 $\pm$ 1.3 & 70.17\\
    \midrule
    \multirow{7}{*}{\rotatebox{90}{\textit{\fontsize{8pt}{8pt}\selectfont Contrastive}}}
    & Infomax &     68.8 $\pm$ 0.8 &   69.9 $\pm$ 3.0 &   75.3 $\pm$ 2.5 &   76.0 $\pm$ 0.7 &   75.9 $\pm$ 1.6 &   58.4 $\pm$ 0.8 &   75.3 $\pm$ 0.5 &   62.7 $\pm$ 0.4 &   64.1 $\pm$ 1.5 &  69.60 \\
    & GraphCL &     69.68 $\pm$ 0.67 & 75.99 $\pm$ 2.65 & 69.80 $\pm$ 2.66 & 78.47 $\pm$ 1.22 & 75.38 $\pm$ 1.44 & 60.53 $\pm$ 0.88 & 73.87 $\pm$ 0.66 & 62.40 $\pm$ 0.57 &  67.88 $\pm$ 0.85 & 70.44 \\
    & JOAO &        70.22 $\pm$ 0.98 & \textbf{81.32} $\pm$ 2.49 & 71.66 $\pm$ 1.43 & 76.73 $\pm$ 1.23 & 77.34 $\pm$ 0.48 & 59.97 $\pm$ 0.79 & 74.98 $\pm$ 0.29 & 62.94 $\pm$ 0.48 & 64.43 $\pm$ 1.38 & 71.07 \\
    & JOAOv2 &      71.39 $\pm$ 0.92 & 80.97 $\pm$ 1.64 & 73.67 $\pm$ 1.00 & 77.51 $\pm$ 1.17 & 75.49 $\pm$ 1.27 & 60.49 $\pm$ 0.74 & 74.27 $\pm$ 0.62 & 63.16 $\pm$ 0.45 & 63.94 $\pm$ 1.59 & 71.21 \\
    & GraphLoG &    72.5 $\pm$ 0.8 & 76.7 $\pm$ 3.3 & 76.0 $\pm$ 1.1 & 77.8 $\pm$ 0.8 & 83.5 $\pm$ 1.2 & \textbf{61.2} $\pm$ 1.1 & 75.7 $\pm$ 0.5 & 63.5 $\pm$ 0.7 & 66.92 $\pm$ 1.58 & 72.65 \\
    & BGRL &        66.73 $\pm$ 1.70 & 64.74 $\pm$ 6.46 & 69.36 $\pm$ 2.66 & 75.52 $\pm$ 1.85 & 71.27 $\pm$ 5.48 & 60.41 $\pm$ 1.43 & 74.83 $\pm$ 0.74 & 63.20 $\pm$ 0.76 & 64.69 $\pm$ 1.66 & 67.86 \\
    \midrule
    
    & D-SLA~(Ours)       & \textbf{72.60} $\pm$ 0.79 & 80.17 $\pm$ 1.50 & \textbf{76.64} $\pm$ 0.91 & \textbf{78.59} $\pm$ 0.44 & \textbf{83.81} $\pm$ 1.01 & 60.22 $\pm$ 1.13 & \textbf{76.81} $\pm$ 0.52 & \textbf{64.24} $\pm$ 0.50 & \textbf{71.56} $\pm$ 0.46 & \textbf{73.85} \\
    \bottomrule
    \end{tabular}
    \end{adjustbox}
    \vspace{-0.1in}
    \label{tab:graph_classification}
\end{table*}

\begin{figure*}[t!]
    \begin{minipage}{0.23\linewidth}
        \vspace{-0.01in}
        \centering
        \begin{adjustbox}{width=\linewidth}
        \renewcommand{\arraystretch}{0.7}
        \begin{tabular}{lcc}
            \toprule
            Dataset & Act. Sim. & Inact. Sim. \\
            \midrule
            BACE    & \textbf{0.6743} & \textbf{0.5403} \\
            \midrule
            HIV     & 0.4186 & 0.4536 \\
            MUV     & 0.1946 & 0.4181 \\
            Tox21   & 0.3047 & 0.3462 \\
            ToxCast & 0.2193 & 0.2962 \\
            SIDER   & 0.2880 & 0.2316 \\
            ClinTox & 0.2725 & 0.2278 \\
            BBBP    & 0.3961 & 0.2031 \\
            \bottomrule
        \end{tabular}
        \end{adjustbox}
        \vspace{-0.08in}
        \captionof{table}{\small Tanimoto similarity on chemical benchmark datasets.}
        \label{tab:analysis}
    \end{minipage}
    \hfill
    \begin{minipage}{0.75\linewidth}
        \centering
        \includegraphics[width=0.99\linewidth]{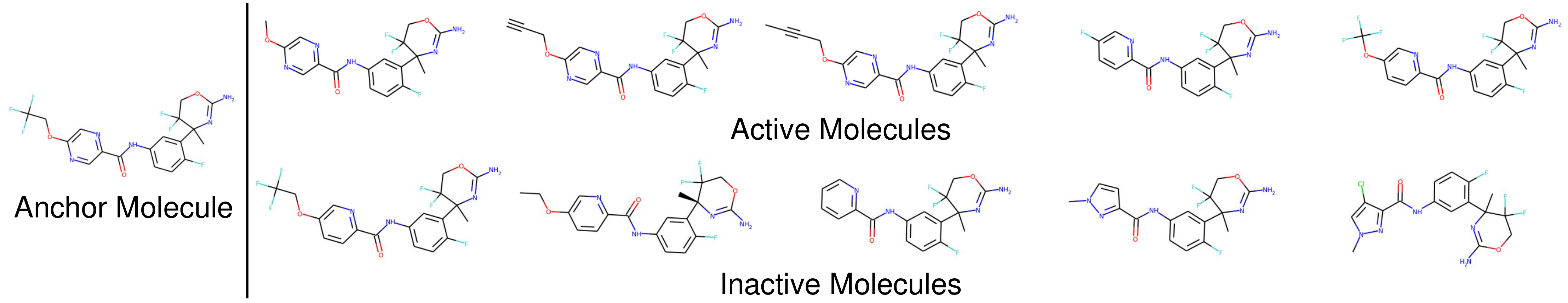}
        \vspace{-0.08in}
        \caption{\small Visualization of molecules in BACE, where the active and inactive molecules in the top and bottom sides are the most similar ones to the molecule in the left, according to the Tanimoto similarity.
        }
        \label{fig:bace}
    \end{minipage}
\vspace{-0.15in}
\end{figure*}
%%%%%%%%%%%%%%%%%%%%%%%%%%%%%%%%%%%%%%%%%%%%%%%%%%%%%%%%%%%%%%%%%%%%%%%%%%%%%%%%%%%%%%s

\vspace{-0.05in}
\paragraph{Analysis}
To analyze whether the structural similarity of molecules is correlated to their biochemical activities, we measure the inherent discrepancy of graphs with the Tanimoto similarity over Morgan fingerprints~\cite{fingerprint}. To be specific, we first iteratively sample an anchor molecule among active molecules in the dataset, and then measure the average Tanimoto similarities of the five most similar active/inactive molecules. In other words, the high similarity values of Inact. Sim. in Table~\ref{tab:analysis} suggests that the molecules have highly overlapped structures regardless of their biochemical activities. For example, as shown in Table~\ref{tab:analysis}, the molecules in the BACE dataset are highly correlated, although their activities are different. Also, we further observe that, as shown in Figure~\ref{fig:bace} that visualizes the most similar active/inactive molecules with respect to the certain anchor molecule in the BACE dataset, the structures between active and inactive molecules are highly similar. 

From the above observations, we suggest that, due to the discrete nature of graphs, two slightly different graphs can have completely different properties, which may be the reason for the performance degeneration of contrastive learning methods in Table~\ref{tab:graph_classification} -- which consider perturbed graphs as similar while they might not be the semantically same -- on such particular datasets (e.g., BACE, MUV, Tox21, ToxCast) having the high Inact. Sim. scores in Table~\ref{tab:analysis}. However, our D-SLA largely outperforms contrastive learning baselines on them\footnote{GraphLoG also adopts the predictive learning scheme by matching the locally masked nodes/subgraphs to their original substructures, thus it shows decent performances on such high similarity datasets.}, because ours not only discriminates an original graph from its perturbations but also can learn their exact discrepancy via the graph edit distance. We note that our method shows the competitive performance on ClinTox and SIDER in Table~\ref{tab:graph_classification}, since they have the lowest structural similarities across different biochemical properties, for which contrastive learning could be effective. We provide more analysis in Appendix \ref{appen:dataset_analysis}.

\subsection{Link Prediction}
\vspace{-0.05in}
Accurately capturing the local semantics of a graph is an important requisite for solving node/edge-level tasks. Thus, we further validate our D-SLA on the link prediction task.

\vspace{-0.05in}
\paragraph{Experimental Setup}
We conduct link prediction experiments on social network datasets --  COLLAB, IMDB-B, and IMDB-M -- from the TU benchmarks~\cite{TUdataset}. We separate the dataset into four parts: pre-training, training, validation, and test sets in the ratio of 5:1:1:3. We use the average precision as an evaluation metric, and report the results over five different runs. We provide more details in Appendix \ref{appen:link_prediction}.

\vspace{-0.05in}
\paragraph{Implementation Details}
For GNNs, we use the GCN~\cite{GCN} following~\citet{JOAO}. For perturbation, we add or delete only a tiny amount of edges (e.g., 1 or 2 edges) while increasing the magnitude of perturbation to obtain three perturbed graphs. Note that we do not use the margin triplet loss in Section~\ref{subsec:margin}, since for local prediction tasks, the graph-level discrepancy learning between completely different graphs is not much helpful in capturing local semantics. For fine-tuning, we train the GNNs to predict whether there is an edge between nodes, and evaluate the GNNs for predicting the existence of 10 edges. For more implementation details, please see Appendix \ref{appen:link_prediction}.

%%%%%%%%%%%%%%%%%%%%%%%%%%%%%%%%%%%%%%%%%%%%%%%%%%%%%%%%%%%%%%%%%%%%%%%%%%%%%%%%%%%%%%
\begin{figure*}[t!]
    \begin{minipage}{0.42\linewidth}
        \centering
        \begin{adjustbox}{width=\columnwidth}
        \renewcommand{\arraystretch}{0.9}
        \begin{tabular}{llccca}
        \toprule
        & SSL Method & COLLAB & IMDB-B & IMDB-M & Avg. \\
        \midrule
        & No Pretrain  & 80.01 $\pm$ 1.14 & 68.72 $\pm$ 2.58 & 64.93 $\pm$ 1.92 & 71.22 \\
        \midrule
        \multirow{2}{*}{\rotatebox{90}{\textit{\footnotesize Pred}}}
        & AttrMasking  & 81.43 $\pm$ 0.80 & 70.62 $\pm$ 3.68 & 63.37 $\pm$ 2.15 & 71.81 \\
        & ContextPred  & 83.96 $\pm$ 0.75 & 70.47 $\pm$ 2.24 & 66.09 $\pm$ 2.74 & 73.51 \\
        \midrule
        \multirow{6}{*}{\rotatebox{90}{\textit{\footnotesize Contra}}}
        & Infomax      & 80.83 $\pm$ 0.62 & 67.25 $\pm$ 1.87 & 64.98 $\pm$ 2.47 & 71.02 \\
        & GraphCL      & 76.04 $\pm$ 1.04 & 63.71 $\pm$ 2.98 & 62.40 $\pm$ 3.04 & 67.38 \\
        & JOAO         & 76.57 $\pm$ 1.54 & 65.37 $\pm$ 3.23 & 62.76 $\pm$ 1.52 & 68.23 \\
        & GraphLoG     & 82.95 $\pm$ 0.98 & 69.71 $\pm$ 3.18 & 64.88 $\pm$ 1.87 & 72.51 \\
        & BGRL         & 76.79 $\pm$ 1.13 & 67.97 $\pm$ 4.14 & 63.71 $\pm$ 2.09 & 69.49 \\
        \midrule
        & D-SLA~(Ours) & \textbf{86.21} $\pm$ 0.38 & \textbf{78.54} $\pm$ 2.79 & \textbf{69.45} $\pm$ 2.29 & \textbf{78.07}\\
        \bottomrule
        \end{tabular}
        \end{adjustbox}
        \vspace{-0.05in}
        \captionof{table}{\small Link prediction results on the social network datasets. The reported results are the average precision, and the bold number denotes the best performance on each dataset.}
        \vspace{-0.075in}
    \label{tab:link_pred}
    \end{minipage}
    \hfill
    \begin{minipage}{0.56\linewidth}
        \centering
        \begin{adjustbox}{width=\columnwidth}
        \begin{minipage}{0.5\linewidth}
        \centerline{\includegraphics[width=0.975\linewidth]{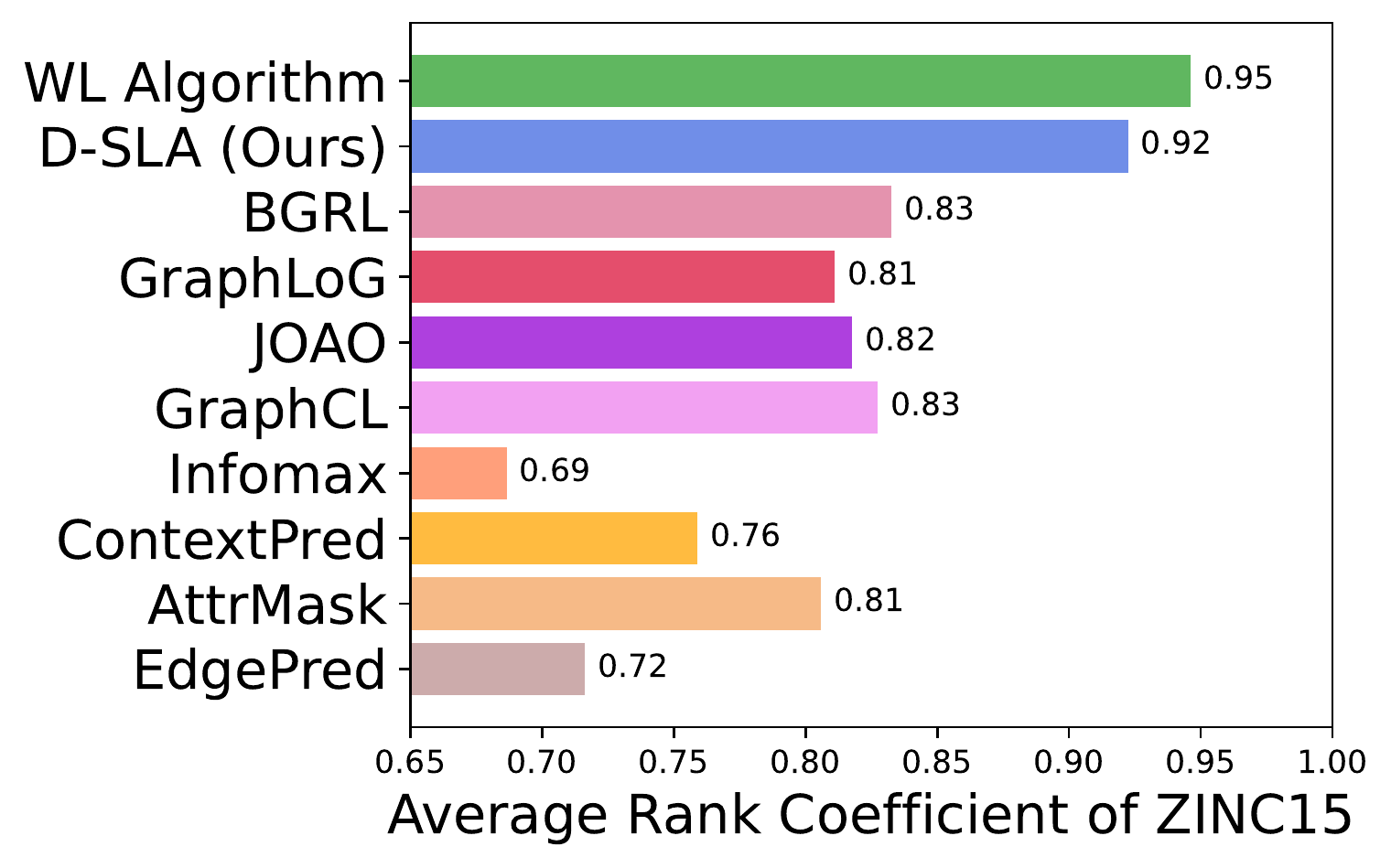}}
        \vspace{-0.05in}
        \end{minipage}
        \hspace{-0.15in}
        \begin{minipage}{0.5\linewidth}
        \centerline{\includegraphics[width=0.975\linewidth]{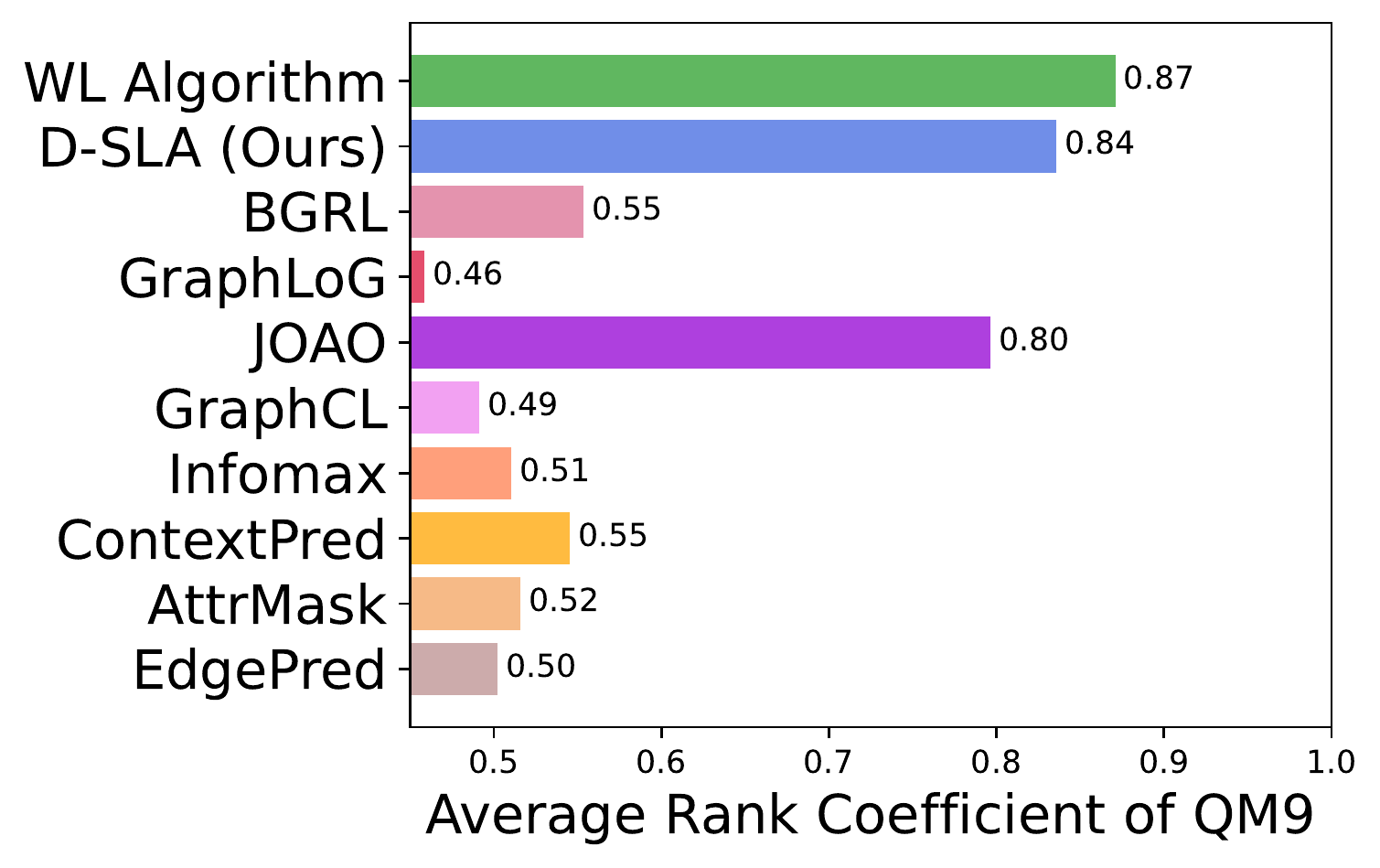}}
        \vspace{-0.05in}
        \end{minipage}
        \end{adjustbox}
        \vspace{-0.05in}
        \captionof{figure}{\small Rank correlation coefficient of 1,000 graphs from ZINC15 and QM9 datasets, measuring the coefficient between the actual similarity ranks and the calculated similarity ranks, between original and perturbed graphs.}
        \vspace{-0.075in}
    \label{fig:rank_coeff}
    \end{minipage}
    \vspace{-0.1in}
\end{figure*}
%%%%%%%%%%%%%%%%%%%%%%%%%%%%%%%%%%%%%%%%%%%%%%%%%%%%%%%%%%%%%%%%%%%%%%%%%%%%%%%%%%%%%%

\vspace{-0.05in}
\paragraph{Results}
As shown in Table~\ref{tab:link_pred}, predictive learning baselines outperform other baselines, since it learns to predict local node/edge attributes. However, our D-SLA largely outperforms all baselines, while it aims to discriminate graph-level representations though. We suggest that this is because ours can capture subtle differences of graphs by leveraging the graph edit distance, demonstrating that accurate discrepancy learning is obviously useful for the local link prediction task.

\subsection{Analysis}
\vspace{-0.05in}
In this section, we further analyze the efficacy of our D-SLA. We provide the additional experimental details in Appendix \ref{appen:analysis}.

\label{subsec:analysis}

\vspace{-0.05in}
\paragraph{Rank Correlation Coefficient}
To see whether learned representations capture the exact amount of discrepancy, we compare the ranks of the calculated vs actual similarities between original and perturbed graphs with Spearman's rank correlation coefficient. Note that the results of the WL algorithm~\cite{WLtest} are merely a performance indicator of discriminative power since it cannot obtain representations that generalize to downstream tasks. As shown in Figure~\ref{fig:rank_coeff}, unlike baselines that mostly fail to discriminate different graphs, our D-SLA has the discriminative power that is on par with the powerful WL algorithm, while ours can generalize to downstream tasks as shown in Table~\ref{tab:graph_classification}.

\vspace{-0.05in}
\begin{wraptable}{t}{0.3\textwidth}
        \centering
        \vspace{-0.175in}
        \captionof{table}{
        \small Ablation study for our D-SLA on graph classification and link prediction tasks.}
        \vspace{-0.05in}
        \begin{adjustbox}{width=\linewidth}
        \renewcommand{\arraystretch}{1.03}
        \renewcommand{\tabcolsep}{0.9mm}
        \begin{tabular}{ccc|cc|c}
        \toprule
        $\mathcal{L}_{GD}$ & $\mathcal{L}_{edit}$ & $\mathcal{L}_{margin}$ & ClinTox & BACE & COLLAB \\
        \midrule
                   &            &            & 58.00 & 70.10 & 71.21 \\
        \checkmark &            &            & 70.83 & 81.58 & 74.23 \\
                   & \checkmark &            & 57.46 & 69.99 & 72.61 \\
        \checkmark & \checkmark &            & 77.48 & 83.53 & 76.19 \\
        \checkmark & \checkmark & \checkmark & 80.17 & 83.81 & N/A \\
        \bottomrule
        \end{tabular}
        \end{adjustbox}
        \vspace{-0.1in}
        
        \label{tab:ablation}

\end{wraptable}

\paragraph{Ablation Study}
To see how much each component contributes to the performance gain, we conduct an ablation study. As shown in Table~\ref{tab:ablation}, we observe that our graph discrimination task $(\mathcal{L}_{GD})$ significantly improves the performance on down-stream tasks against no-pretraining. However, leveraging the exact edit distance alone ($\mathcal{L}_{edit}$) does not learn meaningful representations, since the model trivially sets all the distances between original and its perturbations as zero, as discussed in Section~\ref{subsec:edit}. Also, we further observe that each component of distance-based learning in Section~\ref{subsec:edit} and~\ref{subsec:margin} helps to improve the performance, verifying that accurate discrepancy learning with edit and relative distances is important for modeling graphs. We further discuss the ablation study in Appendix \ref{appen:ablation}.

\vspace{-0.05in}
\paragraph{Embedding Visualization}
We visualize the embedding space from various pre-training methods: predictive, contrastive, and ours, in Figure~\ref{fig:concept} (a,b,c-2). We observe that  predictive learning cannot capture the global graph-level similarity well, as it aims to predict subgraphical semantics of graphs during the pretext task. While contrastive learning can closely embed the highly similar graphs to the original graph, it cannot accurately capture the exact amount of discrepancies among perturbed graphs. Contrarily, our D-SLA can accurately distinguish between the original, perturbed, and negative graphs, while accurately capturing the exact amount of discrepancy for the perturbed graphs.
\section{Conclusion}
In this work, we focused on the limitations of existing self-supervised learning for GNNs: predictive learning does not capture the graph-level similarities; contrastive learning might treat two semantically different graphs from perturbations as similar. To overcome such limitations, we proposed a novel framework (D-SLA) that can learn the graph-level differences among different graphs while also can learn the slight edge-wise differences, by discriminating the original from perturbed graphs. Further, the model is trained to differentiate the target graph from its perturbations and other graphs, while preserving the accurate graph edit distance, allowing the model to discriminate between not only two structurally different graphs but also similar graphs with slight differences. We validated our D-SLA on 12 benchmark datasets, achieving the best average performance. Further analysis shows that it learns a discriminative space of graphs, reflecting the graph edit distances between them. 

\section{Limitation and Potential Societal Impacts}
\label{appen:limitation}
\vspace{-0.05in}
In this section, we discuss the limitation and potential societal impacts of our work.

\paragraph{Limitation}
In this work, we propose a graph self-supervised learning framework, which aims to learn the discrepancy between the original and perturbed graphs, by discriminating the original graph among the original and its perturbations, but also by learning the exact amount of discrepancy across them with the graph edit distance. Note that, while our perturbation scheme can easily make the slight structural differences, it cannot be aware of the semantic differences across differently perturbed graphs, since the graph semantics depend on the target domain. In other words, nodes and edges could differently contribute to the graph semantics depending on the target graph domain (e.g., molecular graphs or social networks), which are hard to pre-define in one framework. As manually annotating the semantic differences across different graphs is impossible to do in an end-to-end fashion, one may need to devise a clever way to be aware of the semantics of graphs for further reflecting the exact semantic difference in the embedding space, which we leave as future work.

\paragraph{Potential Societal Impacts}
Discovering de novo drugs is significantly important to our society since they can be used for curing a disease or enhancing agricultural production, that is directly related to our life. However, it is extremely costly and time-consuming to design drug candidates and validate them, requiring immense lab experiments and labor. In this work, we verify our proposed method in the graph classification task in Table \ref{tab:graph_classification} of the main paper, and ours outperforms the other baselines for classifying toxicity on the Tox21 and ToxCast datasets. We strongly believe that such aspects of our method can positively contribute to our society in developing valuable drugs. However, when validating the toxicity of the most acceptable drug candidates, someone might badly use our method to reduce the money and time (i.e., does not check the toxicity of drugs for reducing costs, however, which may be used for clinical trials to humans), which can negatively affect our society. We sincerely hope that our method would not be utilized for such a bad purpose.

\section{Acknowledgements and Disclosure of Funding}
We thank anonymous reviewers for their constructive feedback. This work was supported by Institute of Information \& communications Technology Planning \& Evaluation (IITP) grant funded by the Korea government (MSIT) (No.2019-0-00075, Artificial Intelligence Graduate School Program (KAIST), and No.2021-0-02068, Artificial Intelligence Innovation Hub), the Engineering Research Center Program through the National Research Foundation of Korea (NRF) funded by the Korean Government MSIT (NRF-2018R1A5A1059921), and Samsung Electronics (IO201214-08145-01).

\bibliography{reference}

%%%%%%%%%%%%%%%%%%%%%%%%%%%%%%%%%%%%%%%%%%%%%%%%%%%%%%%%%%%%

\newpage
\appendix
\vspace{0.3in}
\begin{center}{\bf {\LARGE Appendix \\
}}\end{center}
\vspace{0.4in}

\paragraph{Organization} 
In Section \ref{appen:details}, we first introduce the baselines and our model and then describe the experimental details of graph classification and link prediction tasks but also our in-depth analyses. Then, in Section \ref{appen:additional_results}, we provide the additional experimental results about analyses on datasets, ablation study for our proposed objectives, effects of our hyperparameters ($\lambda_1$, $\alpha$, $\lambda_2$, and the perturbation magnitude), ablation study of attribute masking, and the comparison with augmentation-free approaches.

\section{Experimental Details}
\label{appen:details}
\vspace{-0.05in}
In this section, we first introduce the computing resources that we use, the baselines, and our model in Section~\ref{appen:baselines}. After that, we describe the experimental setups of the graph classification and link prediction tasks in Section~\ref{appen:graph_classification} and Section~\ref{appen:link_prediction} as well as the analysis in Section~\ref{appen:analysis}.

\paragraph{Computing Resources} For all experiments, we use PyTorch and PyTorch Geometric libraries~\cite{pytorch, pytorchgeo}, for easy usage of GPU resources. We use TITAN XP and GeForce RTX 2080 Ti for training and evaluating all models.

\subsection{Baselines and Our Model}
\label{appen:baselines}

\begin{enumerate}[itemsep=2.0mm, parsep=0pt, leftmargin=*]
    \item \textbf{EdgePred} is a predictive learning baseline adopted from the link prediction task of \citet{GraphSAGE}, whose goal is to predict the existence of edges between the given two nodes.
    
    \item \textbf{AttrMasking}~\cite{pretraingnns} is a predictive learning baseline that predicts the attributes of masked nodes and edges from the embeddings of nodes.
    
    \item \textbf{ContxtPred}~\cite{pretraingnns} is a predictive learning baseline that first samples two different subgraphs from the same centered node, and then trains them to be similar while the subgraphs from the other graphs are trained to be dissimilar. 
    
    \item \textbf{Infomax}~\cite{DGI} is a contrastive learning baseline, whose goal is to learn the representations for the given graph and the substructure within the same given graph to be similar while learning the representations for the given graph and the substructures from the negative graphs to be dissimilar.
    
    \item \textbf{GraphCL}~\cite{GraphCL} is a contrastive learning baseline, whose goal is to learn the similarity between two perturbed graphs from the same graph contrasting to in-batch negative graphs over the global graph-level representations. In particular, this method uses the following four perturbation methods: attribute masking, edge perturbing, node dropping, and subgraph sampling.
    
    \item \textbf{JOAO}~\cite{JOAO} is a contrastive learning baseline that, while the learning objective of it is the same as the GraphCL model described above, learns to automatically select the perturbation schemes.
    
    \item \textbf{JOAOv2}~\cite{JOAO} is a variant of JOAO, which has individual projection heads according to the perturbation schemes. Specifically, a perturbed graph is fed into the typical projection head according to the selected perturbation.
    
    \item \textbf{GraphLoG}~\cite{GraphLoG} is a baseline that has two learning objectives: 1) it matches the masked nodes/graphs to their unmasked counterparts; 2) it clusters a group of globally similar graphs with learnable cluster prototypes.
    
    \item \textbf{BGRL}~\cite{BGRL} is a baseline that maximizes the similarity between two perturbed graphs from the original graph without considering in-batch negative graphs, aiming to represent large-scale graphs with efficiency in memory usage.
    
    \item \textbf{D-SLA} is our discrepancy-based graph self-supervised learning framework, which aims to learn the accurate discrepancy between original, perturbed, and negative graphs, by not only discriminating the original graph from its perturbations but also preserving the accurate amount of discrepancy with the graph edit distance between them.

\end{enumerate}

\subsection{Graph Classification}
\label{appen:graph_classification}

%%%%%%%%%%%%%%%%%%%%%%%%%%%%%%%%%%%%%%%%%%%%%%%%%%%%%%%%%%%%%%%%%%%%%%%%%%%%%%%%%%%%%%%%%%%
\begin{wraptable}{t}{0.45\textwidth}
    \centering
    \vspace{-0.17in}
    \caption{\small Dataset statistics on chemical and biological domains.}
    \vspace{-0.1in}
    \begin{adjustbox}{width=\linewidth}
    \begin{tabular}{lcccc}
        \toprule
        Dataset & Tasks & Graphs & Avg. Nodes & Avg. Edges \\
        \midrule
        \multicolumn{5}{l}{\textit{Chemical Domain}} \\
        ZINC15 (Pre-training)  & -   & 2,000,000 & 26.62 & 28.86 \\
        QM9 (Rank Coeff.)   & -   & 133,149  & 8.80  & 9.40 \\
        BBBP                & 1   & 2,039  & 24.06 & 25.95 \\
        ClinTox             & 2   & 1,478  & 26.16 & 27.88 \\
        MUV                 & 17  & 93,087 & 24.23 & 26.28 \\ 
        HIV                 & 1   & 41,127 & 25.51 & 27.47 \\
        BACE                & 1   & 1,513  & 34.09 & 36.86 \\
        SIDER               & 27  & 1,427  & 33.64 & 35.36 \\
        Tox21               & 12  & 7,831  & 18.57 & 19.29 \\
        ToxCast             & 617 & 8,575  & 18.78 & 19.26 \\
        
        \midrule
        \multicolumn{5}{l}{\textit{Biological Domain}} \\
        PPI (Pre-training)     & -  & 306,925 & 39.83 & 364.82 \\
        PPI (Fine-tune)     & 40 & 88,000 & 49.35 &  445.39 \\
        
        \bottomrule
    \end{tabular}
    \end{adjustbox}
    \label{tab:data_stat}
    \vspace{-0.2in}
\end{wraptable}
%%%%%%%%%%%%%%%%%%%%%%%%%%%%%%%%%%%%%%%%%%%%%%%%%%%%%%%%%%%%%%%%%%%%%%%%%%%%%%%%%%%%%%%%%%%
\vspace{-0.075in}
\paragraph{Datasets}
We use the available benchmark datasets\footnote{http://snap.stanford.edu/gnn-pretrain/data/} for the graph classification task. Specifically, for the chemical domain, we use 2M molecules sampled from the ZINC15 dataset~\cite{ZINC15} without using any labels on it. The fine-tuning datasets consist of the molecular graphs from MoleculeNet~\cite{moleculenet}, where the classes are given by the biophysical and physiological properties of the molecules. For the biological domain, the datasets are constructed by the sampled ego-networks from the PPI networks~\cite{PPI/pretrain}. In particular, the pre-training dataset consists of 306K unlabeled protein ego-networks of 50 species, and the fine-tuning dataset consists of 88K protein ego-networks of 8 species with the label given by the functionality of the ego protein. We report the statistics of graph classification datasets in Table~\ref{tab:data_stat}.

\vspace{-0.075in}
\paragraph{Strategy for Selecting Edges for Perturbations}
In this paragraph, we describe the detailed edge selection scheme for our graph perturbation. In our experiments of graph classification, we first select the node and then sample the $3$-hop subgraph of it. After that, we randomly add and remove edges on the subgraph. The reason behind selecting the target subgraph for perturbation is that we aim to reduce the potential risk of making unreasonable cycles, which are impractical especially on the chemical domain. Therefore, to prevent the model to learn such an incorrect bias in the embedding space, we rather sample the subgraph for perturbing the edges.

\vspace{-0.075in}
\paragraph{Common Implementation Details}
We follow the conventional design choice of GNNs for evaluating the graph self-supervised learning methods from \citet{pretraingnns}: Graph Isomorphism Networks (GINs)~\cite{GIN} consisting of 5 layers with 300 dimensions along with mean average pooling for obtaining the entire graph representations. For pre-training of our D-SLA, we sample a subgraph by randomly selecting a center node and then select 3-hop neighbors of it, and then remove the edges on the selected subgraph three times with different magnitudes (20\%, 40\%, 60\%) to make three perturbed graphs, while memorizing the number of deleted and added edges to calculate the graph edit distance. To prevent the situation where the deleted edges are added again, we add edges that are not present in the given original graphs. We mask 80\% of nodes in the selected subgraph to confuse the model to distinguish the original graph from its perturbed graphs. Furthermore, we include the strong perturbation, where 80\% of edges are perturbed and 80\% of nodes are masked among entire nodes and edges in the given graph. $\lambda_1$ and $\lambda_2$ are set as 0.7 and 0.5, respectively.

\vspace{-0.075in}
\paragraph{Implementation Details on Molecular Property Prediction}
We follow the conventional molecule representation setting from~\citet{pretraingnns}, where the node attributes contain the atom number along with the chirality, and the edge attributes contain the bond type (e.g., Single, Double, Triple or Aromatic) along with the bond direction which is represented if an edge is a double or aromatic bond. When adding an edge during edge perturbation, we sample its type by following the distribution of edge attributes in the pre-training dataset. Specifically, we first sample the bond type following the distribution and then sample also the bond direction depending on the bone type. For pre-training, we use the batch size of 256, the number of epochs of 100, the learning rate in the range of [0.01, 0.001, 0.0001], and the margin $\alpha$ in the range of [3,4,5,6,7] by grid search. For the splitting of fine-tuning datasets, we use the scaffold splitting following the conventional setting from~\citet{pretraingnns} and \citet{JOAO}. For fine-tuning, we also follow the conventional setting from \citet{JOAO}.

\vspace{-0.075in}
\paragraph{Implementation Details on Protein Function Prediction}
We use the pre-defined biological graphs from \citet{pretraingnns}, where a node corresponds to a protein without any attributes, and an edge corresponds to a relation type between two proteins such as biological interaction or co-expression. As in molecular property prediction, we add reasonable edges by following the distribution of edge attributes in the pre-training dataset. For pre-training, the number of epochs is 100, the batch size is 128, the learning rate is 0.001, and the margin is 10. For data splitting of the fine-tuning dataset, we use the provided conventional setting from~\citet{pretraingnns}. For fine-tuning, we also follow the conventional setting from \citet{pretraingnns}. Note that, as the result of GraphLoG~\cite{GraphLoG} on this protein function prediction task is not available in the referred paper, we produce the result by following the experimental setups along with the provided public source code.

\vspace{-0.05in}
\subsection{Link Prediction}
\label{appen:link_prediction}
%%%%%%%%%%%%%%%%%%%%%%%%%%%%%%%%%%%%%%%%%%%%%%%%%%%%%%%%%%%%%%%%%%%%%%%%%%%%%%%%%%%%%%%%%%%
\begin{wraptable}{t}{0.4\textwidth}
    \centering
    \vspace{-0.17in}
    \caption{\small Statistics of social network datasets used in link prediction experiments.}
    \vspace{-0.1in}
    \begin{adjustbox}{width=\linewidth}
    \renewcommand{\arraystretch}{1.2}
    \begin{tabular}{lcccc}
        \toprule
        Dataset             & Graphs & Avg. Nodes & Avg. Edges & Pert. Strength \\
        \midrule
        COLLAB              & 4320 & 76.12 & 2331.37 & 0.1\% \\
        IMDB-B              & 2039 & 20.13 & 85.48   & 1\% \\
        IMDB-M              & 1478 & 16.64 & 77.90   & 1\% \\
        \bottomrule
    \end{tabular}
    \end{adjustbox}
    \label{tab:link_data_stat}
    \vspace{-0.15in}
\end{wraptable}        
%%%%%%%%%%%%%%%%%%%%%%%%%%%%%%%%%%%%%%%%%%%%%%%%%%%%%%%%%%%%%%%%%%%%%%%%%%%%%%%%%%%%%%%%%%%

\vspace{-0.05in}
\paragraph{Datasets}
The datasets\footnote{https://chrsmrrs.github.io/datasets/docs/datasets/} we used for the link prediction task are COLLAB, IMDB-B (IMDB-BINARY), IMDB-M (IMDB-MULTI) -- the social network datasets from TU dataset benchmark~\cite{TUdataset}. COLLAB dataset consists of ego-networks extracted from public scientific collaboration networks, namely High Energy Physics, Condensed Matter Physics, and Astro Physics. IMDB-B and IMDB-M are movie collaboration ego-networks where a node represents an actor/actress. The statistics of social network datasets are provided in Table~\ref{tab:link_data_stat}. 

\vspace{-0.05in}
\paragraph{Strategy for Selecting Edges for Perturbations}
To capture the fine-grained local semantics, we suggest that the weaker magnitude of perturbation is the better (See Section~\ref{appen:perturbation_magnitude} verifying the effect of edge perturbation strengths). Therefore, we only perturb the tiny amount of edges (e.g., 1 or 2 edges), as shown in Table~\ref{tab:link_data_stat}, rightmost column.

\vspace{-0.05in}
\paragraph{Implementation Details}
We use the Graph Convolutional Network (GCN)~\cite{GCN} consisting of three layers with 300 hidden dimensions. Following the previous works~\cite{GraphCL, JOAO}, we let the node attributes correspond to the degree of the node. For pre-training, we remove the complete graphs -- that always have the edges between any two nodes -- as we cannot include additional edges during perturbation. For node masking used in AttrMaksing and our D-SLA, we replace the node attribute with the masked token. For hyperparameters, we use the learning rate of 0.001, the batch size of 32, and the $\lambda_1$ of 0.7. During pre-training of our D-SLA, we generate three perturbed graphs by increasing the perturbation magnitudes (e.g., 1\%, 2\%, 3\%). Also, we further mask 20\% of nodes in perturbation. 

\vspace{-0.05in}
\subsection{Analysis}
\label{appen:analysis}
\vspace{-0.05in}
\paragraph{Rank Correlation Coefficient}
The spearman's rank correlation coefficient measures the correlation between two rank series in the range from -1 to 1, where the value is 1 if the two rank series are perfectly and monotonically the same. We build the following two rank series to compare: 1) the labeled similarity rank between the original and perturbed graphs using the graph edit distance, and 2) the predicted similarity rank based on the embedding-level distances between original and perturbed graphs from pre-trained models. Specifically, we perturb edges of the entire graph by gradually increasing the magnitude of edge perturbation (i.e., 5\%, 10\%, 15\%, 25\%, 35\%, 45\%, 60\%, 75\%, 90\%), and then label ranks of the perturbed graphs to the original graph according to the graph edit distance. Then, the original and perturbed graphs are fed into the pre-trained models, and the ranks are measured by the embedding-level distance between the original and perturbed graphs. Therefore, if a pre-trained model can capture the exact amount of discrepancy, the rank correlation coefficient would be 1, by locating the embedding of a similar graph (a weakly perturbed graph) closer to the original graph than the embedding of a dissimilar graph (a strongly perturbed graph). We measure the coefficient with randomly sampled $1,000$ different graphs.

For models to calculate the similarity across different graphs, we use the pre-trained model for the graph classification task in our D-SLA. For EdgePred, AttrMasking, ContextPred, Informax, and GraphCL baselines, we use the publicly available pre-trained models\footnote{https://github.com/snap-stanford/pretrain-gnns, https://github.com/Shen-Lab/GraphCL}. For JOAO and GraphLoG, we use the public source codes\footnote{https://github.com/Shen-Lab/GraphCL\_Automated, https://github.com/DeepGraphLearning/GraphLoG}, to obtain the pre-trained models. The WL algorithm in Figure 5 of the main paper corresponds to a randomly initialized GIN model. In other words, since the GIN is as powerful as the WL test, we denote it as the WL algorithm. We evaluate the above models on two different datasets: ZINC15~\cite{ZINC15} and QM9~\cite{TUdataset}, where statistics of each dataset is provided in Table~\ref{tab:data_stat}. Note that the QM9 dataset is not used for pre-training, thus we can measure the model's generalization ability with it.

\vspace{-0.05in}
\paragraph{Embedding Visualization}
To visualize the representation space of the original, perturbed, and negative graphs, we pre-train the models on the subset of the ZINC15 dataset~\cite{ZINC15} with the perturbation strategy in Section \ref{appen:graph_classification} for obtaining perturbed graphs. Then, after pre-training, we visualize the graph representations by PCA~\cite{pca} and t-SNE~\cite{tsne} for Figure \ref{fig:concept} and Figure \ref{fig:emb_ablation} of the main paper, respectively.

%%%%%%%%%%%%%%%%%%%%%%%%%%%%%%%%%%%%%%%%%%%%%%%%%%%%%%%%%%%%%%%%%%%%%%%%%%%%%%%%%%%%%%%%%%%
\begin{figure*}[t]
    \centering
    \begin{subfigure}[b]{0.78\textwidth}
        \centering
        \includegraphics[width=\textwidth]{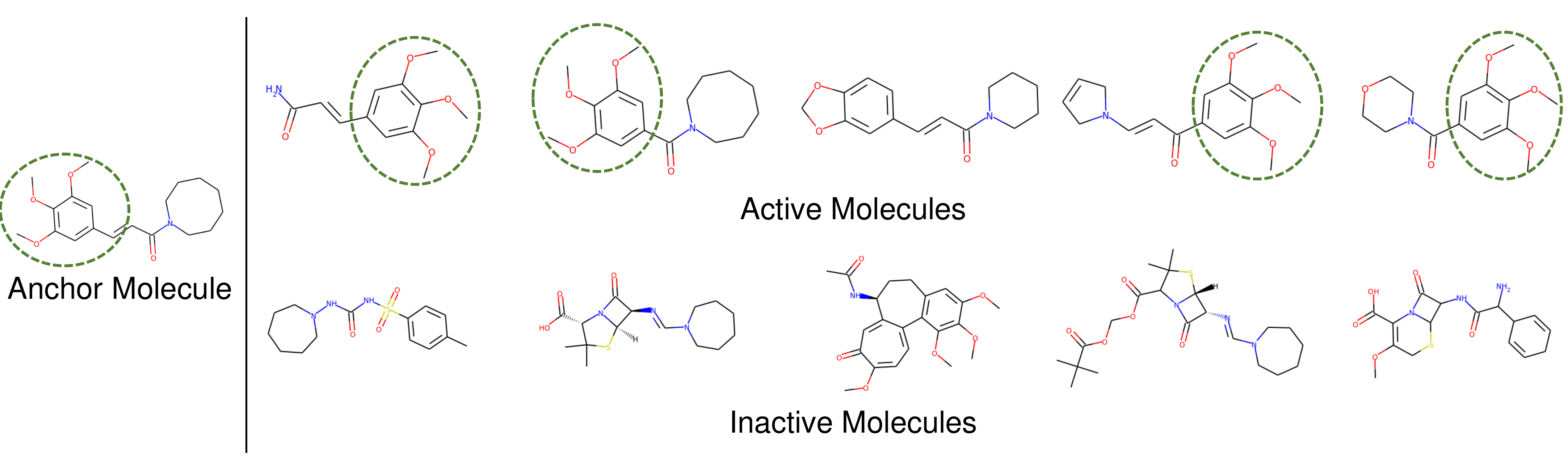}
        \vspace{-0.25in}
        \subcaption{BBBP}
    \end{subfigure}
    \begin{subfigure}[b]{0.78\textwidth}
        \centering
        \includegraphics[width=\textwidth]{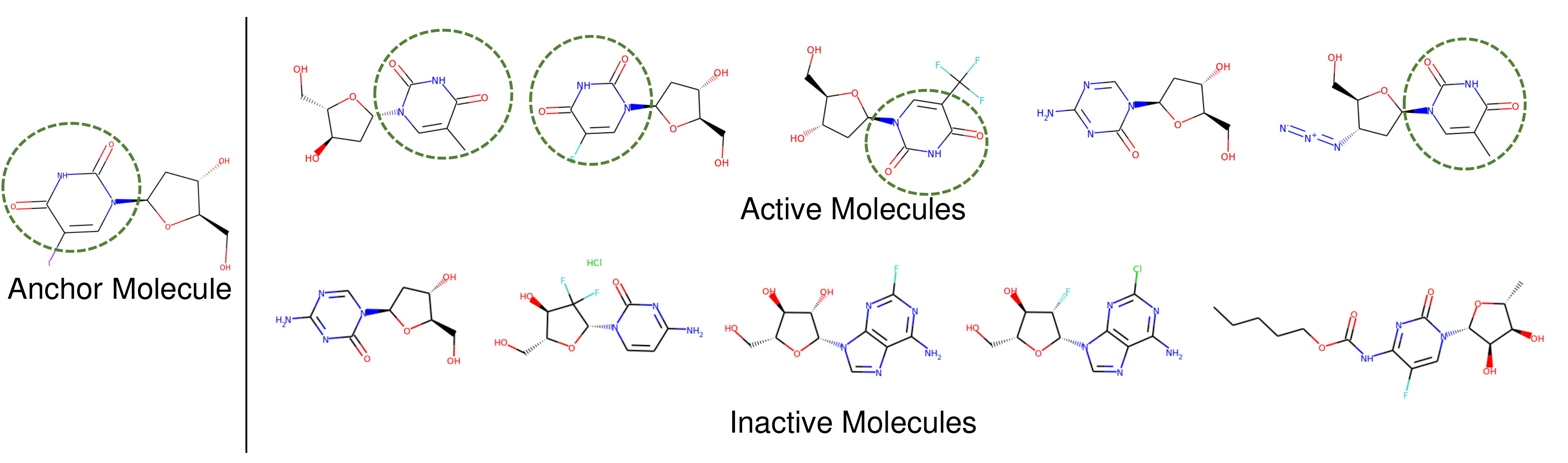}
        \vspace{-0.25in}
        \subcaption{ClinTox}
    \end{subfigure}
    \begin{subfigure}[b]{0.78\textwidth}
        \centering
        \includegraphics[width=\textwidth]{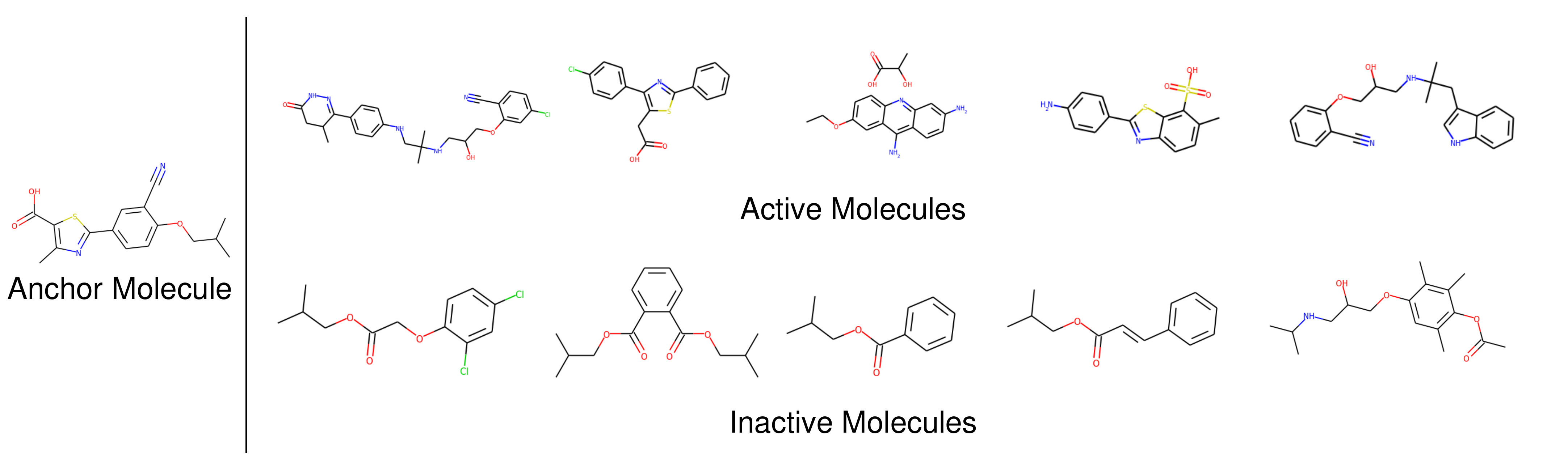}
        \vspace{-0.25in}
        \subcaption{Tox21}
    \end{subfigure}
    \vspace{-0.1in}
    \captionof{figure}{\small Structural comparisons on the most similar (i.e., top-5) active/inactive molecules for the certain anchor active molecule on the left side, for the BBBP, ClinTox, and Tox21 datasets. Green dotted circles indicate the shared structure across different molecules.}
    \label{fig:appen_analysis}
    \vspace{-0.2in}
\end{figure*}
%%%%%%%%%%%%%%%%%%%%%%%%%%%%%%%%%%%%%%%%%%%%%%%%%%%%%%%%%%%%%%%%%%%%%%%%%%%%%%%%%%%%%%%%%%%

\vspace{-0.05in}
\section{Additional Experimental Results}
\vspace{-0.05in}
\label{appen:additional_results}
In this section, we provide additional results with their corresponding discussions. To be specific, in Section~\ref{appen:dataset_analysis}, we analyze the correlation between the characteristics of the dataset and self-supervised learning methods. Then, we provide an in-depth discussion about our observations in the ablation study in Section~\ref{appen:ablation}. Additionally, we provide some guidelines for choosing our hyperparameters ($\lambda_1$, $\alpha$, $\lambda_2$, and the perturbation magnitude) in Section \ref{appen:lambda_1}, \ref{appen:alpha}, \ref{appen:lambda_2}, and \ref{appen:perturbation_magnitude}, respectively. Futhermore, we provide an ablation study of attribute masking in Section~\ref{appen:ablation_attr_masking}. Finally, we compare our D-SLA with augmentation-free approaches in Section~\ref{appen:augmentation-free}.

\vspace{-0.05in}
\subsection{Dataset Analysis}
\label{appen:dataset_analysis}
\vspace{-0.05in}
In this subsection, we further discuss the characteristics of graph self-supervised learning methods with respect to the characteristics of datasets. As shown in Section \ref{subsec:graph_classification}, we find that contrastive learning methods outperform predictive learning methods on BBBP and ClinTox. Contrarily, predictive learning methods outperform contrastive learning methods on Tox21. Therefore, we further analyze BBBP, ClinTox, and Tox21 datasets to answer why such methods have counterfactual effects on different datasets.

In Figure~\ref{fig:appen_analysis}, we visualize the structures of active/inactive molecules from the anchor molecule. We observe that in BBBP and ClinTox datasets, the activities are highly correlated to the structural similarity, i.e., the structurally similar molecules show the same activities. Therefore, as contrastive learning aims to maximize the similarity between perturbed graphs from the original graph, it fits into the BBBP and ClinTox datasets, showing better performance than predictive learning methods. However, in the Tox21 dataset, we cannot observe any clues that the activities are correlated to the structural similarity. Therefore, capturing the structural similarity with contrastive learning seems to be useless in this dataset, resulting in the better performance of predictive learning methods. However, our D-SLA can learn the discrete embedding space by learning the discrepancy even between similar graphs, thus obtaining a discriminative space that can further be utilized to distinguish between them for downstream tasks and outperforming all other baselines as shown in Section \ref{subsec:graph_classification}.

\vspace{-0.05in}
\subsection{Additional Interpretation of Ablation Study}
\label{appen:ablation}
\vspace{-0.05in}
We conduct an ablation study on link prediction and graph classification tasks in Table \ref{tab:ablation} of the main paper. For link prediction, we observe that both two components, $\mathcal{L}_{GD}$ and $\mathcal{L}_{edit}$, consistently improve the performance, thus verifying that our discrepancy-based learning allows the model to capture the local semantics of graphs. Also, for graph classification, we choose the most different two datasets in their properties -- ClinTox and BACE datasets -- to obviously see the contribution of each component in our D-SLA. In particular, for the ClinTox dataset in which the biochemical activities of molecules are highly correlated to their structures, we observe that it is important to discriminate the negative graphs from the perturbed graphs with the triplet margin loss $\mathcal{L}_{margin}$, as the performance improvements on using it is significant compared to the other dataset: BACE. However, in the case of the BACE dataset, since the molecules are highly similar regardless of their biochemical activities, the graph edit distance loss $\mathcal{L}_{edit}$ largely contributes to the performance gain, allowing the model to learn the exact discrepancy across similar graphs. 

%%%%%%%%%%%%%%%%%%%%%%%%%%%%%%%%%%%%%%%%%%%%%%%%%%%%%%%%%%%%%%%%%%%%%%%%%%%%%%%%%%%%%%%%%
\begin{figure*}[t]
    \begin{minipage}{0.24\textwidth}
    \centering
    \centerline{\includegraphics[width=0.975\linewidth]{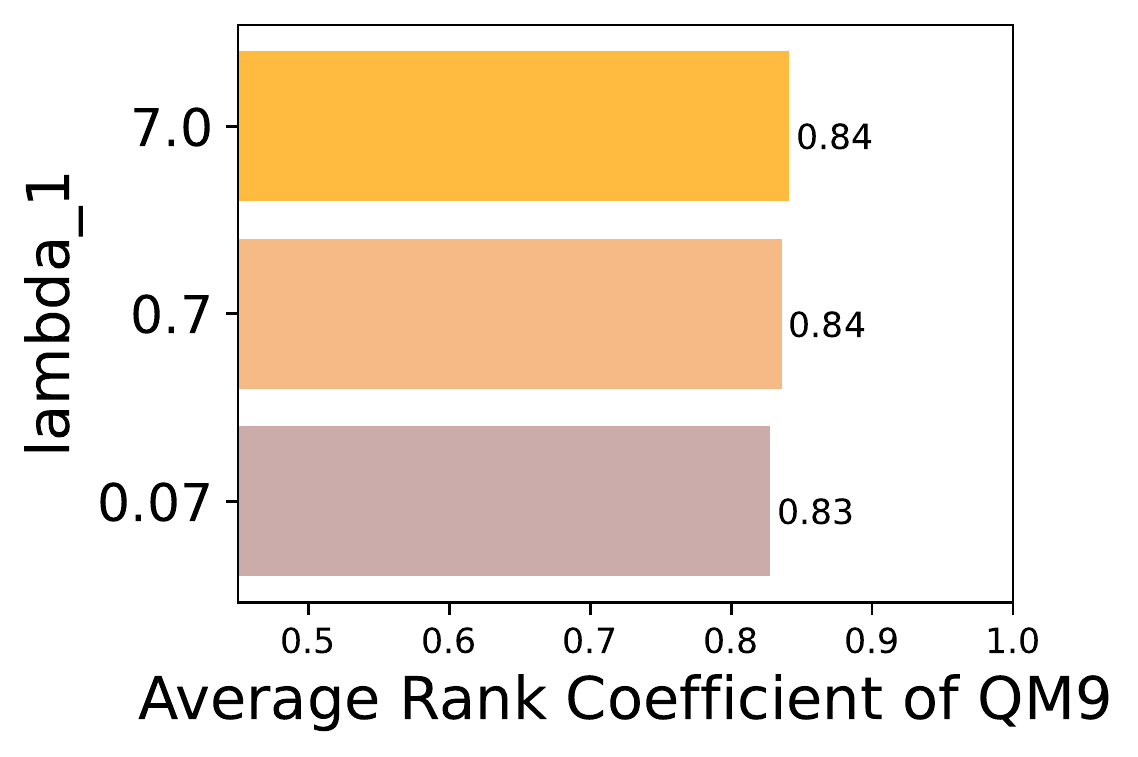}}
    \vspace{-0.05in}
    \caption{\small Rank correlation coefficient of QM9 with varying $\lambda_1$ values.}
    \label{fig:lambda1_rank_coeff}
    \end{minipage}
    \hfil
    \begin{minipage}{0.74\textwidth}
    \centering
        \begin{minipage}{0.31\linewidth}
        \centering
        \centerline{\includegraphics[width=0.975\linewidth]{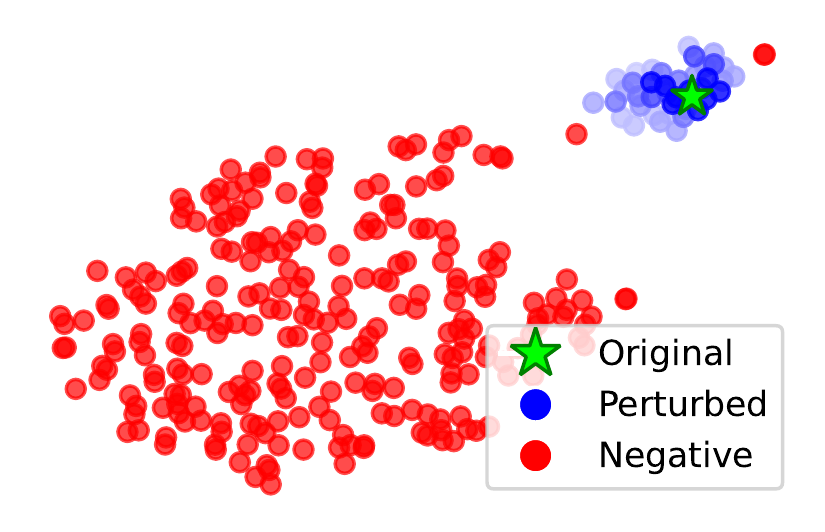}}
        \vspace{-0.1in}
        \subcaption{$\lambda_1=0.07$}
        \end{minipage}
        \hfill
        \begin{minipage}{0.31\linewidth}
        \centering
        \centerline{\includegraphics[width=0.975\linewidth]{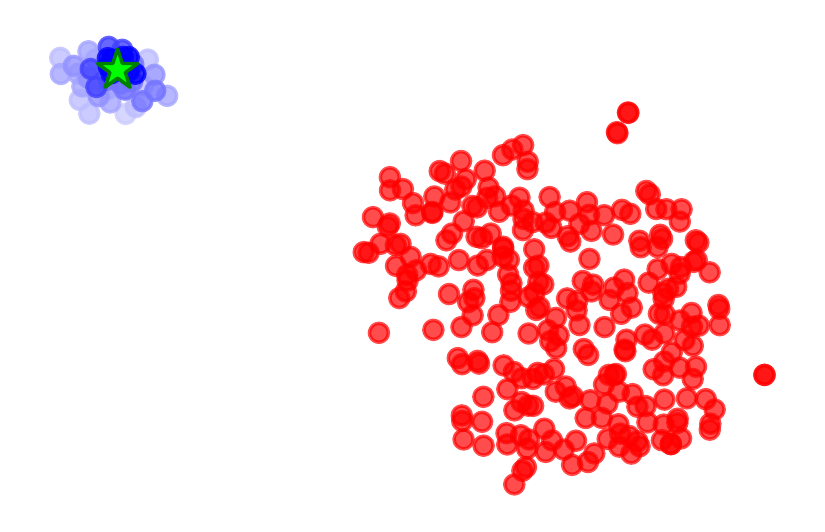}}
        \vspace{-0.1in}
        \subcaption{$\lambda_1=0.7$}
        \end{minipage}
        \hfill
        \begin{minipage}{0.31\linewidth}
        \centering
        \centerline{\includegraphics[width=0.975\linewidth]{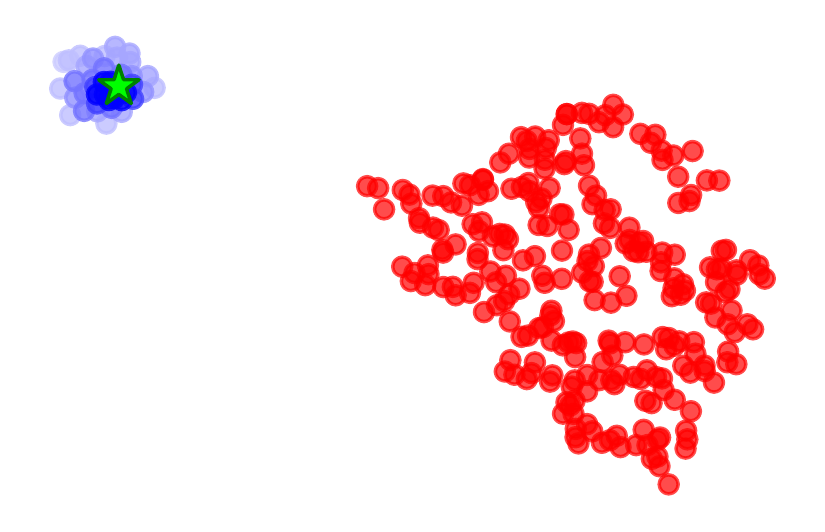}}
        \vspace{-0.1in}
        \subcaption{$\lambda_1=7.0$}
        \end{minipage}
        \vspace{-0.08in}
        \caption{\small Visualization of the learned latent representation space for different $\lambda_1$ values. Note that each model is trained on the subset of the ZINC15 dataset and the embedding spaces are visualized by t-SNE~\cite{tsne}.}
        \label{fig:lambda1_emb_vis}
    \end{minipage}
    \vspace{-0.1in}
\end{figure*}
%%%%%%%%%%%%%%%%%%%%%%%%%%%%%%%%%%%%%%%%%%%%%%%%%%%%%%%%%%%%%%%%%%%%%%%%%%%%%%%%%%%%%%%%%

\vspace{-0.05in}
\subsection{Effect of $\mathcal{L}_{edit}$ Coefficient ($\lambda_1$)}
\label{appen:lambda_1}
\vspace{-0.05in}
We demonstrate the effect of the coefficient $\lambda_1$ for $\mathcal{L}_{edit}$ (Equation \ref{eq:overall}). Specifically, we show its efficacy by measuring the rank correlation coefficient, and by visualizing the graph representation space. Firstly, as shown in Figure~\ref{fig:lambda1_rank_coeff}, the differences in rank correlation with varying scaling coefficient $\lambda_1$ are marginal. Also, as shown in Figure~\ref{fig:lambda1_emb_vis}, for all models, the perturbed graphs are well embedded along their perturbation magnitudes regardless of $\lambda_1$ values. Therefore, we suggest that, fortunately, learning the accurate amount of discrepancy between graphs does not heavily depend on the scaling hyperparameter for $\mathcal{L}_{edit}$, namely $\lambda_1$. On the other hand, we observe that, if $\lambda_1$ is relatively small, some negative graphs are embedded closer to the perturbed ones (Figure~\ref{fig:lambda1_emb_vis} (a)). This result indicates that, the regularization effects of our objective $\mathcal{L}_{edit}$ for learning the discrepancy between the original and perturbed graphs can also affect the boundary between perturbed and negative graphs, as shown in Figure~\ref{fig:lambda1_emb_vis} (b), (c).

%%%%%%%%%%%%%%%%%%%%%%%%%%%%%%%%%%%%%%%%%%%%%%%%%%%%%%%%%%%%%%%%%%%%%%%%%%%%%%%%
\begin{wraptable}{t}{0.20\textwidth}
    \vspace{-0.17in}
    \centering
    \captionof{table}{\small Effect of $\alpha$ on BACE dataset.}
    \vspace{-0.05in}
    \begin{adjustbox}{width=1.0\linewidth}
        \renewcommand{\arraystretch}{1.0}
        \begin{tabular}{lc}
            \toprule
            $\alpha$ & ROC-AUC \\
            \midrule
            1.0  & 83.75 $\pm$ 0.96 \\
            5.0  & 83.81 $\pm$ 1.01 \\
            10.0 & 78.34 $\pm$ 1.07 \\
            \bottomrule
        \end{tabular}
    \end{adjustbox}
    \vspace{-0.1in}
    \label{tab:alpha}
\end{wraptable}
%%%%%%%%%%%%%%%%%%%%%%%%%%%%%%%%%%%%%%%%%%%%%%%%%%%%%%%%%%%%%%%%%%%%%%%%%%%%%%%%%%%%%%%%%%%%%%%%

\vspace{-0.05in}
\subsection{Effect of Margin ($\alpha$) in $\mathcal{L}_{margin}$}
\label{appen:alpha}
\vspace{-0.05in}
To demonstrate the effect of the margin $\alpha$ in $\mathcal{L}_{margin}$ (Equation \ref{eq:margin}), we pre-train the model with varying $\alpha$ values and fine-tune it on BACE dataset for graph classification. As shown in Table \ref{tab:alpha}, large value of $\alpha$ degenerates our discrepancy learning among similar graphs. As described in Section \ref{subsec:margin}, if the distance between the original and its negative graph ($d'_j$) is larger than the distance between the original and perturbed graph ($d_i$) plus $\alpha$ (i.e., $\alpha + d_i < d'_j$), the distance between the original and perturbed graph ($d_i$) is preserved not losing the discrepancy learned by $\mathcal{L}_{edit}$ (Equation \ref{eq:editdistance}). However, too large $\alpha$ makes it hard to satisfy the condition (i.e., $\alpha + d_i < d'_j$) forcing the model to lose the discrepancy learned by $\mathcal{L}_{edit}$.

%%%%%%%%%%%%%%%%%%%%%%%%%%%%%%%%%%%%%%%%%%%%%%%%%%%%%%%%%%%%%%%%%%%%%%%%%%%%%%%%
\begin{wraptable}{t}{0.20\textwidth}
    \vspace{-0.17in}
    \centering
    \captionof{table}{\small Effect of $\lambda_2$ on BACE dataset.}
    \vspace{-0.05in}
    \begin{adjustbox}{width=1.0\linewidth}
        \renewcommand{\arraystretch}{1.0}
        \begin{tabular}{lc}
            \toprule
            $\lambda_2$ & ROC-AUC \\
            \midrule
            0.1 & 83.68 $\pm$ 0.78 \\
            0.5 & 83.81 $\pm$ 1.01 \\
            0.9 & 80.72 $\pm$ 0.71 \\
            \bottomrule
        \end{tabular}
    \end{adjustbox}
    \vspace{-0.1in}
    \label{tab:lambda2}
\end{wraptable}
%%%%%%%%%%%%%%%%%%%%%%%%%%%%%%%%%%%%%%%%%%%%%%%%%%%%%%%%%%%%%%%%%%%%%%%%%%%%%%%%%%%%%%%%%%%%%%%%

\vspace{-0.05in}
\subsection{Effect of $\mathcal{L}_{margin}$ Coefficient ($\lambda_2$)}
\label{appen:lambda_2}
We demonstrate the effect of the coefficient $\lambda_2$ for $\mathcal{L}_{margin}$ (Equation \ref{eq:contra}) by pre-training with various $\lambda_2$ values and fine-tuning on BACE dataset for graph classification. As shown in Table \ref{tab:lambda2}, when $\lambda_2$ is large, the performance on BACE downstream dataset is generated, indicating that the model cannot learn the discrepancy among similar graphs. We suggest that this is because $\lambda_2$ controls the intensity of attracting the similar graphs of $\mathcal{L}_{margin}$ and a large value of $\lambda_2$ which would strongly attract the similar graphs forces the model to lose the discrepancy among similar graphs learned by $\mathcal{L}_{edit}$ (Equation \ref{eq:editdistance}).

%%%%%%%%%%%%%%%%%%%%%%%%%%%%%%%%%%%%%%%%%%%%%%%%%%%%%%%%%%%%%%%%%%%%%%%%%%%%%%%%%%%%%%%%%%%
\begin{wraptable}{t}{0.25\textwidth}
    \centering
    \vspace{-0.17in}
    \caption{\small Effect of magnitude of perturbation on the link prediction task.}
    \vspace{-0.07in}
    \begin{adjustbox}{width=1.0\linewidth}
    \renewcommand{\arraystretch}{1.0}
    \begin{tabular}{lc}
        \toprule
        Magnitude & Accuracy \\
        \midrule
        10\%  & 70.67 $\pm$ 0.63 \\
        1\%   & 74.55 $\pm$ 0.76 \\
        0.1\% & 76.19 $\pm$ 0.50 \\
        \bottomrule
    \end{tabular}
    \end{adjustbox}
    \vspace{-0.2in}
    \label{tab:magnitude}
\end{wraptable}
%%%%%%%%%%%%%%%%%%%%%%%%%%%%%%%%%%%%%%%%%%%%%%%%%%%%%%%%%%%%%%%%%%%%%%%%%%%%%%%%%%%%%%%%%%%

\vspace{-0.05in}
\subsection{Effect of Perturbation Magnitude on Link Prediction}
\label{appen:perturbation_magnitude}
\vspace{-0.05in}
We validate the effect of the perturbation magnitude on the COLLAB dataset for link prediction. As shown in Table~\ref{tab:magnitude}, we observe that the performance of link prediction is enhanced when only a small amount of edges are perturbed, demonstrating that weaker perturbation magnitude is better for capturing local semantics. If the perturbation magnitude is weak, the perturbed graphs are slightly different from the original graph, thus the model could capture a subtle difference across original and perturbed graphs. 

%%%%%%%%%%%%%%%%%%%%%%%%%%%%%%%%%%%%%%%%%%%%%%%%%%%%%%%%%%%%%%%%%%%%%%%%%%%%%%%%%%%%%%%%%%%
\begin{wraptable}{t}{0.27\textwidth}
    \centering
    \vspace{-0.17in}
    \caption{\small Ablation study of attribute masking on the link prediction task.}
    \vspace{-0.07in}
    \begin{adjustbox}{width=1.0\linewidth}
    \renewcommand{\arraystretch}{1.0}
    \begin{tabular}{lc}
        \toprule
         & Accuracy \\
        \midrule
        w/ Masking  & 76.19 $\pm$ 0.50 \\
        w/o Masking   & 70.42 $\pm$ 0.95 \\
        \bottomrule
    \end{tabular}
    \end{adjustbox}
    \vspace{-0.2in}
    \label{tab:attrmask}
\end{wraptable}
%%%%%%%%%%%%%%%%%%%%%%%%%%%%%%%%%%%%%%%%%%%%%%%%%%%%%%%%%%%%%%%%%%%%%%%%%%%%%%%%%%%%%%%%%%%

\subsection{Ablation Study of Attribute Masking}
\label{appen:ablation_attr_masking}
\vspace{-0.05in}
We conduct an additional ablation study for the attribute masking in our perturbation strategy on the COLLAB dataset for link prediction. As shown in Table \ref{tab:attrmask}, the performance without attribute masking is significantly lower than the performance with attribute masking. We suggest that, in the pre-training stage, attribute masking limits the information given to the model and forces the model to learn more transferable and fruitful representations, demonstrating that attribute masking in our perturbation is a key factor to learn the local semantics.

%%%%%%%%%%%%%%%%%%%%%%%%%%%%%%%%%%%%%%%%%%%%%%%%%%%%%%%%%%%%%%%%%%%%%%%%%%%%%%%%%%%%%%%%%%%
\begin{table*}[t]
    \caption{\small Fine-tuning results on graph classification tasks. Best performances are highlighted in bold.
    }
    \vspace{-0.075in}
    \begin{adjustbox}{width=\textwidth}
    \renewcommand{\arraystretch}{1.0}
    \begin{tabular}{lcccccccca}
    \toprule
    SSL methods & BBBP & ClinTox & MUV & HIV & BACE & SIDER & Tox21 & ToxCast & Avg. \\
    \midrule
    SimGCL~\cite{SimGCL} & 67.37 $\pm$ 1.23 &	55.66 $\pm$ 4.72 &	71.24 $\pm$ 1.79 &	75.04 $\pm$ 0.86 &	74.11 $\pm$ 2.74 &	57.44 $\pm$ 1.74 &	74.39 $\pm$ 0.45 &	62.27 $\pm$ 0.38 &	67.19 \\
    SimGRACE~\cite{SimGRACE} &  71.25 $\pm$ 0.86 & 64.16 $\pm$ 4.50 & 71.18 $\pm$ 3.40 & 74.52 $\pm$ 1.12 & 73.81 $\pm$ 1.37 & \textbf{60.59} $\pm$ 0.96 & 74.20 $\pm$ 0.64 & 63.36 $\pm$ 0.52 & 69.13 \\
    \midrule
    D-SLA~(Ours)       & \textbf{72.60} $\pm$ 0.79 & \textbf{80.17} $\pm$ 1.50 & \textbf{76.64} $\pm$ 0.91 & \textbf{78.59} $\pm$ 0.44 & \textbf{83.81} $\pm$ 1.01 & 60.22 $\pm$ 1.13 & \textbf{76.81} $\pm$ 0.52 & \textbf{64.24} $\pm$ 0.50 & \textbf{74.51} \\
    \bottomrule
    \end{tabular}
    \end{adjustbox}
    \vspace{-0.1in}
    \label{tab:aug_classificaiton}
\end{table*}
%%%%%%%%%%%%%%%%%%%%%%%%%%%%%%%%%%%%%%%%%%%%%%%%%%%%%%%%%%%%%%%%%%%%%%%%%%%%%%%%%%%%%%%%%%

%%%%%%%%%%%%%%%%%%%%%%%%%%%%%%%%%%%%%%%%%%%%%%%%%%%%%%%%%%%%%%%%%%%%%%%%%%%%%%%%%%%%%%%%%%%
\begin{figure*}[t]
    \centering
    \begin{minipage}{0.49\textwidth}
    \centering
        \begin{minipage}{0.32\linewidth}
        \centering
            \centerline{\includegraphics[width=0.975\linewidth]{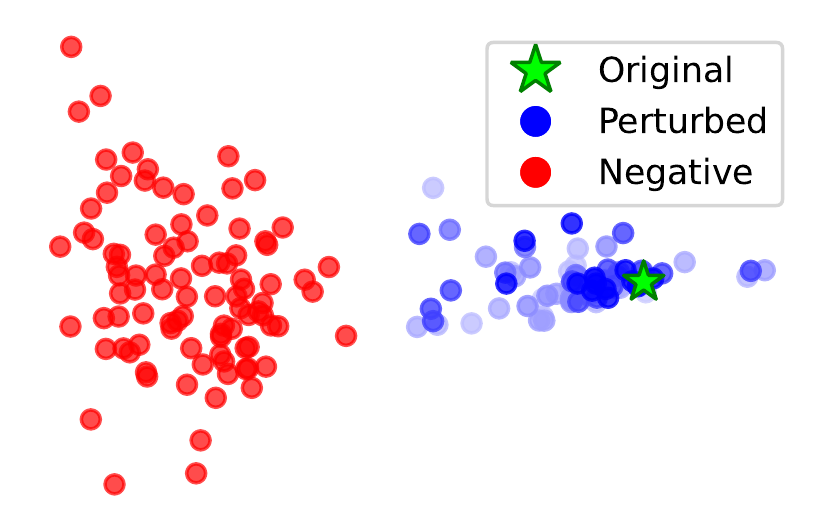}}
            \vspace{-0.1in}
            \subcaption{SimGCL}
        \end{minipage}
        \hfill
        \begin{minipage}{0.32\linewidth}
        \centering
            \centerline{\includegraphics[width=0.975\linewidth]{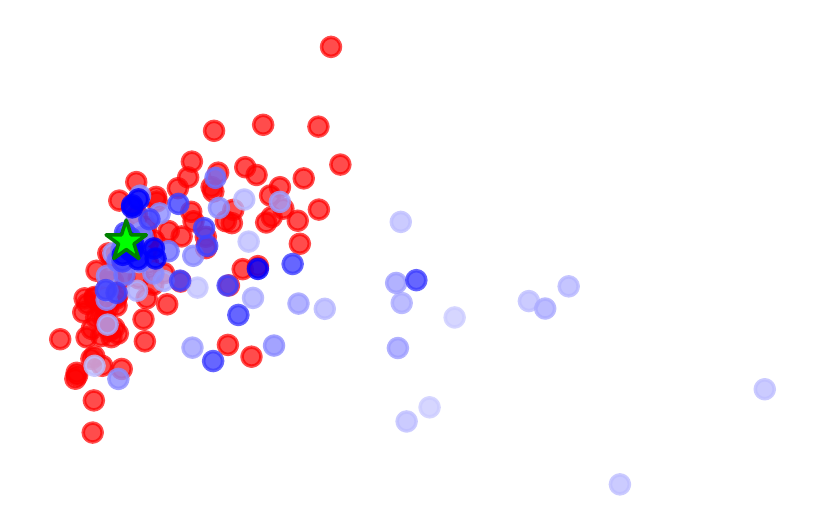}}
            \vspace{-0.1in}
            \subcaption{SimGRACE}
        \end{minipage}
        \hfill
        \begin{minipage}{0.32\linewidth}
        \centering
            \centerline{\includegraphics[width=0.975\linewidth]{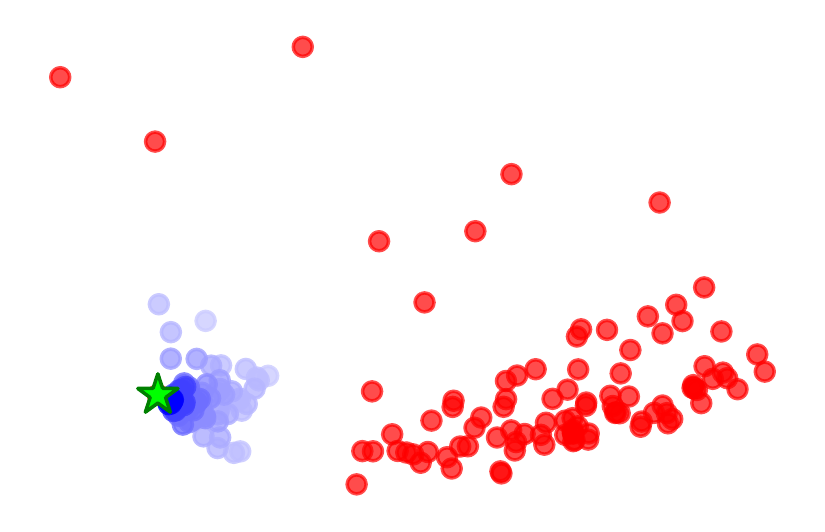}}
            \vspace{-0.1in}
            \subcaption{D-SLA (Ours)}
        \end{minipage}
        \vspace{-0.1in}
        \caption{\small Embedding space visualization on similar and dissimilar graphs with Graph Edit Distance.}
        \label{fig:aug_emb_vis_ged}
    \end{minipage}
    \hfill
    \begin{minipage}{0.49\textwidth}
    \centering
        \begin{minipage}{0.32\linewidth}
        \centering
            \centerline{\includegraphics[width=0.975\linewidth]{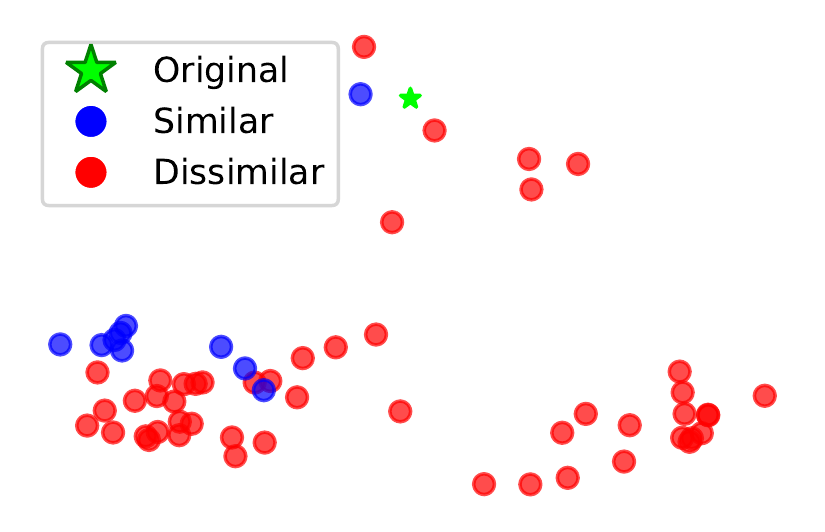}}
            \vspace{-0.1in}
            \subcaption{SimGCL}
        \end{minipage}
        \hfill
        \begin{minipage}{0.32\linewidth}
        \centering
            \centerline{\includegraphics[width=0.975\linewidth]{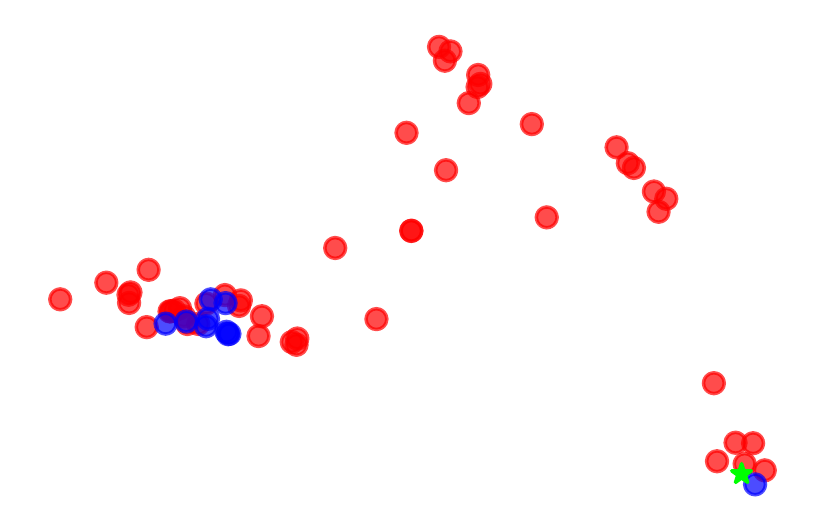}}
            \vspace{-0.1in}
            \subcaption{SimGRACE}
        \end{minipage}
        \hfill
        \begin{minipage}{0.32\linewidth}
        \centering
            \centerline{\includegraphics[width=0.975\linewidth]{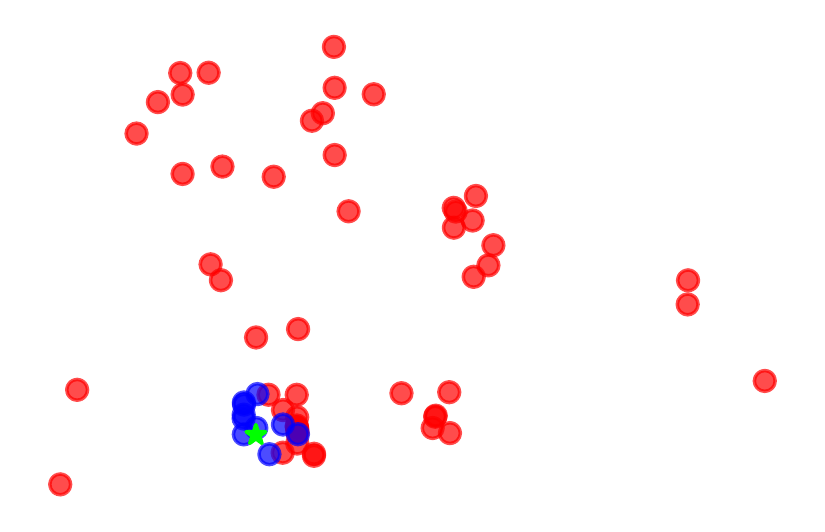}}
            \vspace{-0.1in}
            \subcaption{D-SLA (Ours)}
        \end{minipage}
        \vspace{-0.1in}
        \caption{\small Embedding space visualization on similar and dissimilar graphs with Tanimoto Similarity.}
        \label{fig:aug_emb_vis_tanimoto}
    \end{minipage}
    \vspace{-0.1in}
\end{figure*}
%%%%%%%%%%%%%%%%%%%%%%%%%%%%%%%%%%%%%%%%%%%%%%%%%%%%%%%%%%%%%%%%%%%%%%%%%%%%%%%%%%%%%%%%%%%

%%%%%%%%%%%%%%%%%%%%%%%%%%%%%%%%%%%%%%%%%%%%%%%%%%%%%%%%%%%%%%%%%%%%%%%%%%%%%%%%%%%%%%%%%%
\begin{wraptable}{t}{0.45\textwidth}
    \vspace{-0.175in}
    \caption{\small Fine-tuning results on link prediction tasks. Best performances are highlighted in bold.}
    \vspace{-0.05in}
    \begin{adjustbox}{width=1.0\linewidth}
        \renewcommand{\arraystretch}{1.0}
        \begin{tabular}{lccca}
            \toprule
             & COLLAB & IMDB-B & IMDB-M & Avg. \\
            \midrule
            SimGCL & 77.46 $\pm$ 0.86 & 64.91 $\pm$ 2.60 & 63.78 $\pm$ 2.28 & 68.72 \\
            SimGRACE & 74.51 $\pm$ 1.54 & 64.49 $\pm$ 2.79 & 62.81 $\pm$ 2.32 & 67.27 \\
            \midrule
            D-SLA (Ours) & \textbf{86.21} $\pm$ 0.38 & \textbf{78.54} $\pm$ 2.79 & \textbf{69.45} $\pm$ 2.29 & \textbf{78.07} \\
            \bottomrule
        \end{tabular}
    \end{adjustbox}
    \vspace{-0.1in}
   
    \label{tab:aug_link}
\end{wraptable}
%%%%%%%%%%%%%%%%%%%%%%%%%%%%%%%%%%%%%%%%%%%%%%%%%%%%%%%%%%%%%%%%%%%%%%%%%%%%%%%%%%%%%%%%%%

\subsection{Comparison with Augmentation-Free Contrastive Learning Approaches}
\label{appen:augmentation-free}
\vspace{-0.05in}
Recently, augmentation-free contrastive learning methods have been proposed. We validate the effectiveness of our discrepancy learning framework by comparing with augmentation-free approaches on graph classification and link prediction tasks. We compare our D-SLA with SimGCL~\cite{SimGCL} and SimGRACE~\cite{SimGRACE} that augment views of graphs by adding noise to graph embeddings or model parameters while preserving the graph structures. As shown in Table \ref{tab:aug_classificaiton} and Table \ref{tab:aug_link}, our discrepancy learning outperforms the augmentation-free contrastive learning approaches, demonstrating the effectiveness of our discrepancy learning framework in capturing both local and global semantics.
We further visualize the embedding space of similar and dissimilar graphs with different distance metrics such as Graph Edit Distance (Figure \ref{fig:aug_emb_vis_ged}) and Tanimoto similarity (Figure \ref{fig:aug_emb_vis_tanimoto}). We observe that, in SimGCL and SimGRACE, the similar and dissimilar graphs are not distinguished and the augmentation-free approaches cannot capture the exact amount of discrepancy, since they cannot learn the difference between similar graphs. Contrarily, by our discrepancy learning, the model can distinguish similar and dissimilar graphs and learn the exact amount of discrepancy both on Graph Edit Distance and Tanimoto Similarity.

\end{document}